\documentclass[11pt]{article}
\usepackage{acl}
\usepackage{times}
\usepackage{latexsym}
\usepackage[T1]{fontenc}
\usepackage[utf8]{inputenc}
\usepackage{microtype}
\usepackage{inconsolata}
\usepackage{graphicx}
\usepackage{booktabs}
\usepackage{placeins}
\usepackage{amsmath}
\usepackage{xcolor}
\usepackage{tikz}
\usetikzlibrary{arrows.meta,positioning,fit,backgrounds,calc,shapes.geometric,shadows.blur}
\usepackage{pifont}
\definecolor{cFroz}{RGB}{219,234,254}\definecolor{cFrozB}{RGB}{37,99,235}
\definecolor{cTrain}{RGB}{255,237,213}\definecolor{cTrainB}{RGB}{234,88,12}
\definecolor{cSem}{RGB}{220,242,225}\definecolor{cSemB}{RGB}{22,143,74}
\definecolor{cVis}{RGB}{237,226,250}\definecolor{cVisB}{RGB}{124,58,180}
\newcommand{\yes}{\textcolor{green!55!black}{\ding{51}}}
\newcommand{\no}{\textcolor{red!62!black}{\ding{55}}}
\newcommand{\pt}{\textcolor{black!45}{\footnotesize$\sim$}}
\newcommand{\blfootnote}[1]{%
  \begingroup
  \renewcommand\thefootnote{}\footnote{#1}%
  \addtocounter{footnote}{-1}%
  \endgroup
}
\title{Predict, Then Retrieve: Cross-Instance \\ Future-State Retrieval from Video Prefixes}

\author{Quynh Vo \quad
Thong Nguyen$^{\dagger\,}$ \quad
Vinh-Hien Do \quad
Cong-Duy Nguyen \quad 
Anh-Tuan Luu \quad \\\\
Centre for AI Research, VinUniversity \\
National University of Singapore
\\\\
Email: \texttt{thong.nguyen@u.nus.edu}
}

\begin{document}
\maketitle

\begin{abstract}
We introduce \textbf{Predictive State Retrieval (PSR)}, a task in which a model observes a short video
\emph{prefix} and a temporal question about an object's future state, then retrieves instances from
\emph{other} videos or images that depict that state. Unlike action anticipation, which predicts a label,
moment retrieval, which localizes an observed event within a video, or video generation, which synthesizes
pixels, PSR combines \emph{anticipation} with \emph{cross-instance retrieval} across multiple temporal
horizons. We construct a benchmark from four datasets with graded, human-validated ground truth, difficulty
tiers, and an oracle ceiling. We also propose \textbf{LFTR}, a lightweight retriever with frozen encoders that
predicts a question- and horizon-conditioned future latent and matches it in complementary semantic and visual
spaces. A ceiling decomposition reveals a clear bottleneck: the true future state is highly retrievable once
specified, whereas every predictor we evaluate, including a large multimodal language model with access to the
prefix frames, remains far below the oracle. Thus, \emph{forecasting rather than perception is the central
learnable challenge}. LFTR narrows this gap at substantially lower inference cost, and ablations attribute its
gains to cross-space fusion and hard-negative training rather than latent rollout. We release the benchmark,
code, and evaluation scripts.
\end{abstract}
\blfootnote{$^{\dagger}$Corresponding author.}
\begin{figure*}[t]\centering
\includegraphics[width=\textwidth]{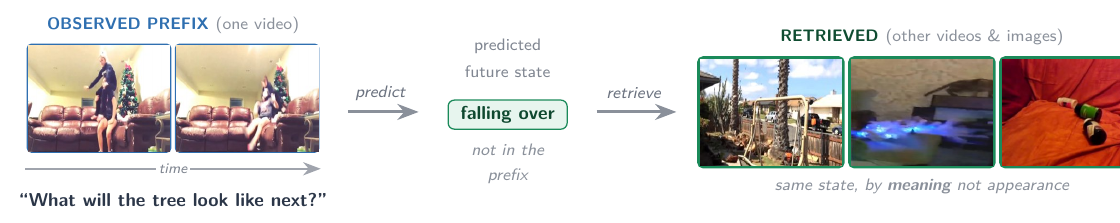}
\caption{\textbf{Predictive State Retrieval (PSR).} From a video \emph{prefix} (blue) and a temporal question,
a model must \emph{predict} the object's future state (gold) and \emph{retrieve} instances from \emph{other}
videos/images (green) that depict it --- the answer is never in the prefix, so retrieval is cross-instance and
matches by \emph{meaning}, not appearance.}
\label{fig:teaser}
\end{figure*}

\section{Introduction}
\label{sec:intro}
Anticipating how a scene will change is central to how people act in the world: seeing a glass tip over the
edge of a table, we expect it to shatter, and can immediately bring to mind \emph{other} scenes that show the
same outcome. Reproducing this ability has driven a large body of video understanding. \emph{Action
anticipation} forecasts the label of a not-yet-observed action \citep{ego4d,epickitchens,furnari2020rulstm,
girdhar2021avt,vondrick2016anticipating}; \emph{world models} roll a scene forward in a latent space for
planning and control \citep{vjepa2,oord2018cpc}; and \emph{video generation} synthesizes future frames
outright, increasingly with retrieval-augmented motion \citep{motionrag}. In parallel, \emph{video--text
retrieval} and \emph{moment localization} let us search large corpora with language
\citep{luo2022clip4clip,xu2021videoclip,wang2024internvideo2,bain2021frozen,vcmr,lei2020tvr,gao2017tall,
lei2021qvhighlights}, and vision--language models now describe and reason over scenes at scale
\citep{radford2021clip,zhai2023siglip,li2023blip2,liu2023llava}.

Yet these threads stop short of the capability the opening example takes for granted: from what has been
\emph{observed so far}, \emph{predict} how an object will look next and \emph{retrieve} concrete examples of
that future state from \emph{other} videos and images. Anticipation predicts a future but collapses it to a
fixed label rather than an open set of instances \citep{ego4d,epickitchens}; moment retrieval searches a
corpus but targets a described moment that has \emph{already occurred}, \emph{within} a single video
\citep{vcmr,lei2020tvr}; composed and cross-instance retrieval match across instances, but from a
\emph{present} query rather than a predicted future \citep{vo2019composing,liu2021cirr,ventura2024covr}; and
world models predict the future but keep it \emph{latent and internal}, never surfacing matching evidence
\citep{vjepa2}. No existing benchmark asks a model to \emph{forecast a state and then retrieve cross-instance
evidence of it} --- and this gap is what we set out to fill.

We introduce \textbf{Predictive State Retrieval (PSR)} (Fig.~\ref{fig:teaser}): from a video \emph{prefix} and
a temporal question about an object, retrieve instances --- from a corpus of \emph{other} videos and images
--- that depict the object's predicted future state. PSR is hard for two reasons that set it apart from
existing tasks. The target is \emph{not observed}, so the model must \emph{predict} a future that grows more
uncertain with the horizon; and the answer lives in \emph{different} videos, so it must be matched by
\emph{meaning}, not appearance --- retrieving frames that merely look like the present will not do. Because
ground truth links instances through their shared semantic \emph{state}, PSR is \emph{language-grounded by
design}: the bridge across instances is linguistic, and we show empirically (Sec.~\ref{sec:analysis}) that raw
appearance cannot recover this ground truth whereas the semantic-state abstraction is highly retrievable.

Our method follows the same two steps a person would --- \emph{anticipate, then recall by meaning}. We propose
\textbf{LFTR}, a \emph{frozen}-encoder retriever that predicts a question- and horizon-conditioned future
latent from the prefix and matches it against the corpus in two complementary spaces: a \emph{semantic} space
for \emph{what} state is expected and a \emph{visual} space for how it looks, fused at the score level.
Building the predictor on frozen encoders keeps it cheap and transparent, and isolates \emph{forecasting} as
the one component that must be learned. That isolation yields our central finding, a \emph{ceiling
decomposition}: the true future state is highly retrievable and nearly horizon-flat, yet no predictor we run
--- including a sighted large multimodal model --- comes close to that ceiling, so \emph{forecasting, not
perception, is the core learnable bottleneck}. LFTR narrows the gap at a fraction of the cost, with its gains
coming from fusing complementary spaces and from hard negatives rather than from rolling the latent forward.

\noindent We make three contributions:
\begin{itemize}\itemsep2pt
\item We introduce \textbf{Predictive State Retrieval (PSR)}, a benchmark that asks a model to \emph{predict}
an object's future state from a short video prefix and \emph{retrieve} cross-instance examples of it from
\emph{other} videos and images, with graded difficulty tiers, an oracle ceiling, and human-validated ground
truth.
\item We propose \textbf{LFTR}, a cheap frozen-encoder retriever that predicts a question- and
horizon-conditioned future latent and retrieves it by \emph{cross-space fusion} of a semantic and a visual
match --- where what helps is \emph{complementarity, not count}.
\item We establish, through a \emph{ceiling decomposition}, that the learnable bottleneck is
\emph{forecasting, not perception}, and turn the method--ceiling gap into a concrete, measurable open problem
--- one our cheap retriever already narrows further than a far larger zero-shot MLLM.
\end{itemize}

\section{Related Work}
\label{sec:related}

\begin{table*}[t]\centering\small\setlength{\tabcolsep}{4pt}\renewcommand{\arraystretch}{0.9}
\caption{\textbf{PSR vs.\ prior settings.} Each prior task covers only part: anticipation predicts the future
but \emph{classifies} a label; retrieval/grounding target an \emph{observed} moment; world models predict a
latent \emph{within} one video. PSR is the only setting that is future-targeted, cross-instance,
retrieval-native, multi-horizon, spans \emph{both} video and image corpora, and is scored against a graded
ground truth with an oracle ceiling. (\yes\,yes\quad\no\,no\quad\pt\,partial; scale figures for prior work are
approximate.)}
\label{tab:compare}
\resizebox{\textwidth}{!}{%
\begin{tabular}{lccccc cc rrc}
\toprule
& \multicolumn{5}{c}{\textbf{Capabilities}} & \multicolumn{2}{c}{\textbf{Media}} & \multicolumn{3}{c}{\textbf{Scale}} \\
\cmidrule(lr){2-6}\cmidrule(lr){7-8}\cmidrule(lr){9-11}
Benchmark & Future & Cross-inst. & Retrieval & Multi-horiz. & Open-vocab & Video & Image & Corpus & \#Queries & \#Q-types \\
\midrule
\multicolumn{11}{l}{\emph{Prediction --- action / state-change anticipation, procedure planning}}\\
Ego4D~\citep{ego4d}                 & \yes & \no  & \no  & \pt  & \no  & \yes & \no  & $\sim$9.6k & $3.5$k  & 1 \\
EPIC-100~\citep{epickitchens}       & \yes & \no  & \no  & \pt  & \no  & \yes & \no  & $700$      & $\sim$90k  & 1 \\
CrossTask~\citep{zhukov2019crosstask}& \yes & \no  & \no  & \pt  & \pt  & \yes & \no  & $4.8$k    & $\sim$2.7k & 1 \\
\midrule
\multicolumn{11}{l}{\emph{Retrieval / grounding}}\\
QVHighlights~\citep{lei2021qvhighlights} & \no  & \no  & \yes & \no  & \yes & \yes & \no  & $10.1$k   & $10.3$k & 1 \\
DiDeMo~\citep{hendricks2017didemo}  & \no  & \no  & \yes & \no  & \yes & \yes & \no  & $10.5$k    & $\sim$40k  & 1 \\
TVR~\citep{lei2020tvr}              & \no  & \no  & \yes & \no  & \yes & \yes & \no  & $21.8$k    & $109$k     & 1 \\
Charades-STA~\citep{gao2017tall}    & \no  & \no  & \yes & \no  & \yes & \yes & \no  & $6.7$k     & $16$k      & 1 \\
MSR-VTT~\citep{xu2016msrvtt}        & \no  & \yes & \yes & \no  & \yes & \yes & \no  & $10$k      & $200$k     & 1 \\
WebVid~\citep{bain2021frozen}       & \no  & \yes & \yes & \no  & \yes & \yes & \no  & $2.5$M     & $2.5$M     & 1 \\
CIRR~\citep{liu2021cirr}            & \no  & \yes & \yes & \no  & \yes & \no  & \yes & $21.6$k    & $36.5$k    & 1 \\
WebVid-CoVR~\citep{ventura2024covr} & \no  & \yes & \yes & \no  & \yes & \yes & \no  & $131$k     & $1.6$M     & 1 \\
\midrule
\multicolumn{11}{l}{\emph{Object states / understanding}}\\
MIT-States~\citep{isola2015states}  & \no  & \pt  & \pt  & \no  & \pt  & \no  & \yes & $63$k     & $63$k   & 1 \\
VAW~\citep{pham2021vaw}             & \no  & \pt  & \pt  & \no  & \pt  & \no  & \yes & $72$k     & $260$k     & 1 \\
ChangeIt~\citep{soucek2022changeit} & \no  & \pt  & \pt  & \no  & \pt  & \yes & \no  & $34$k     & $34$k      & 1 \\
HowToChange~\citep{xue2024howtochange}& \no  & \pt  & \pt  & \no  & \pt  & \yes & \no  & $5.4$k   & $5.4$k     & 1 \\
\midrule
\textbf{PSR (ours)}                 & \yes & \yes & \yes & \yes & \yes & \yes & \yes & \textbf{218k$+$333k} & \textbf{20{,}182} & \textbf{5} \\
\bottomrule
\end{tabular}}
\end{table*}

\subsection{Anticipation, world models, and object states}
\label{sec:rel_anticip}
Work on \emph{predicting the future} spans three threads that PSR draws on but differs from. \emph{Action
anticipation} benchmarks \citep{ego4d,epickitchens} and models \citep{girdhar2021avt,furnari2020rulstm}
forecast the \emph{label} of an upcoming action a few seconds ahead --- closed-vocabulary classification over
a fixed horizon, not open cross-instance retrieval. \emph{Future-representation prediction} ---
anticipating future visual features from unlabeled video \citep{vondrick2016anticipating}, joint-embedding
predictive architectures \citep{vjepa2}, and contrastive predictive coding \citep{oord2018cpc} --- predicts a
\emph{latent} future, but for recognition or planning \emph{within} a video (e.g.\ procedure
planning \citep{niu2024schema}), never as a cross-instance retrieval target with a graded ground truth.
\emph{Object-state} recognition and state-change discovery
\citep{isola2015states,pham2021vaw,xue2024howtochange,soucek2022changeit}
supplies our state vocabulary but treats state as per-image classification or within-video localization. Most
related, \emph{object state-change anticipation} \citep{manousaki2024osca} forecasts an object's forthcoming
state transition --- but \emph{within} a single procedural video and as anticipation/classification, not as
cross-instance retrieval against a corpus.
PSR unifies these into a \emph{predictive, cross-instance retrieval} task across horizons: the target is
unobserved, lives in \emph{other} videos, and is matched by state \emph{meaning}. LFTR borrows the
predict-in-latent idea but conditions on the \emph{question and horizon}, reads into a retrieval space, and
trains with hard-negative contrastive learning \citep{robinson2021hardneg}.

\subsection{Retrieval and vision--language models}
\label{sec:rel_retrieval}
A large body of work learns joint video--text spaces for retrieval
\citep{luo2022clip4clip,xu2021videoclip,bain2021frozen,wang2022internvideo,wang2024internvideo2,zhu2024languagebind},
and video (corpus) moment retrieval \citep{vcmr,lei2021qvhighlights} localizes a described moment
\emph{within} a video; broad image--text models \citep{radford2021clip,zhai2023siglip} and multimodal LLMs
\citep{li2023blip2,alayrac2022flamingo,liu2023llava,lin2023videollava} add strong zero-shot
recognition and captioning. In all of these the query describes an \emph{observed} moment; PSR's query is a
prefix plus a \emph{future} question, so the target has not yet occurred and lives in \emph{other} videos ---
requiring prediction, not just matching. We use the strongest such encoders as both zero-shot baselines and
LFTR target spaces (App.~\ref{app:matrix}), and find native video--text alignment does not, on its own, yield
a better \emph{predictable} target than a self-supervised image space. Closest in form, \emph{composed image
retrieval} \citep{vo2019composing,liu2021cirr,saito2023pic2word} --- and its video counterpart,
\emph{composed video retrieval} \citep{ventura2024covr} --- retrieves a target given a source image/video
plus a \emph{modification text} (``current state $+$ transformation $\rightarrow$ target''), which is PSR
minus the \emph{prediction} of the transformation. This invites the objection that PSR reduces to
(forecast a state caption) $+$ (text$\rightarrow$X retrieval); our multimodal-LLM baseline
(\textbf{M3}, \S\ref{sec:exp}) is exactly that pipeline, and our fair-comparison study
(\S\ref{sec:analysis}, Table~\ref{tab:m3fair}) shows it \emph{underperforms} a model that predicts a
\emph{native visual latent} --- a target no text-caption forecaster can reach with any text encoder.

\noindent\textbf{Closest work and novelty.} PSR is not the first task to \emph{predict} a future (action
anticipation) or to \emph{retrieve} across instances (composed/video retrieval); its novelty is their
\emph{union}, and that the union is not trivial. Concurrent efforts touch parts but none combine the
ingredients: STATUS Bench \citep{ukai2025status} and CAST \citep{liu2026cast} handle object-state
\emph{transitions} and retrieval but from \emph{observed} inputs, not a future forecast; composed video
retrieval \citep{ventura2024covr} matches cross-instance but is \emph{given} the transformation as text
rather than predicting it (our M3 baseline is exactly that pipeline, and loses; \S\ref{sec:exp}). To our
knowledge PSR is thus the first benchmark to unite \emph{future-state prediction} with \emph{cross-instance
corpus retrieval} across horizons, scored against a graded ground truth and an oracle ceiling
(Table~\ref{tab:compare}) --- and its ceiling decomposition (\S\ref{sec:analysis}) shows the union exposes a
genuinely new bottleneck (\emph{forecasting}), not a solved composition.

\section{The PSR Benchmark}
\label{sec:bench}

\begin{figure*}[t]\centering
\includegraphics[width=\textwidth]{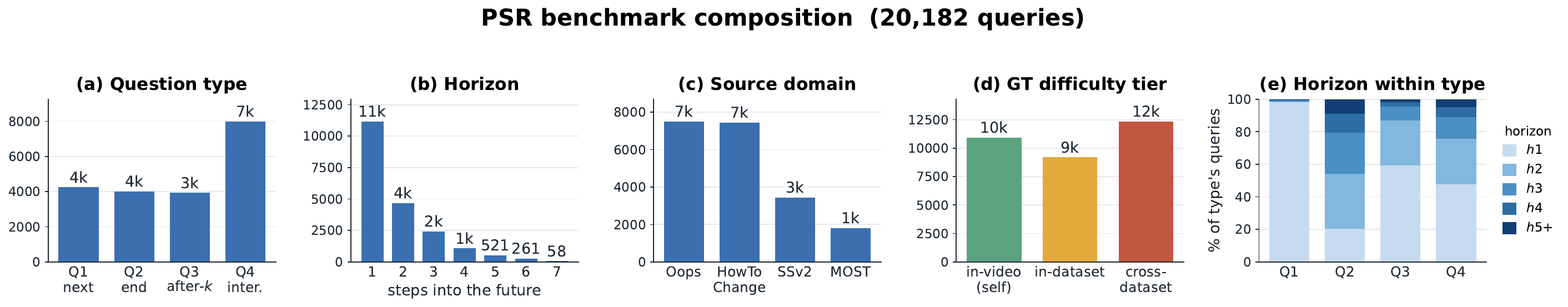}
\caption{\textbf{PSR benchmark composition} over all 20{,}182 queries: (a) question types; (b) horizon;
(c) source domains; (d) difficulty tier (by where the correct instance lives); (e) horizon \emph{within} each
question type. Queries skew to near horizons and the cross-dataset (hardest) tier dominates.}
\label{fig:stats}
\end{figure*}

\subsection{Task and Construction}
\textbf{Task definition.} A query is $q=(x,\tau,h)$ where $x$ is a video prefix, $\tau$ a question type, and
$h$ a temporal horizon. Given a frozen corpus $\mathcal{C}$ of instances (video segments / images), the goal
is to rank $\mathcal{C}$ so that the ground-truth set $\mathcal{P}_q\subset\mathcal{C}$ --- instances
depicting the queried future state --- appears at the top, for \emph{all} question types with \emph{one}
model --- a capability profile no prior benchmark offers (Table~\ref{tab:compare}). We report
Recall@$\{1,5,10,20,50\}$ and MRR. The graded structure of the GT (difficulty tiers,
state granularity, consensus) is exploited by \emph{stratifying} the evaluation (App.~\ref{app:tiers},~\ref{app:gtrobust})
rather than by collapsing it into a single graded score; a tier-weighted nDCG variant is included in the
released evaluation code for users who prefer one number.

\textbf{Question types.} Q1 \emph{next state}; Q2 \emph{end state}; Q3 \emph{state after $k$ seconds};
Q4 \emph{intermediate state} (before reaching a named state); Q5 \emph{multi-object} (the joint future
states of several objects). Each maps to a horizon $h$, so a question is an \emph{index into the future
trajectory} of the object. App.~\ref{app:qtypes} gives the full taxonomy---per-type definitions, worked
examples, query counts, and horizon coverage (Table~\ref{tab:qtypes}).

\begin{figure*}[t]\centering
\includegraphics[width=\textwidth]{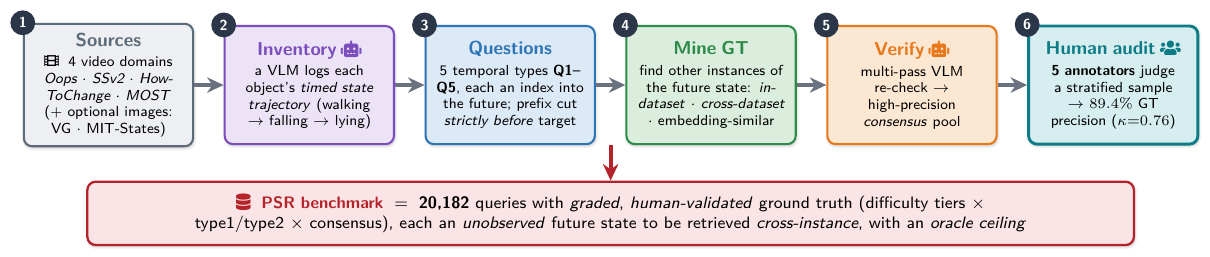}
\caption{\textbf{How the PSR benchmark is built.} A VLM inventories each object's timed state trajectory in
four video sources; we generate five temporal question types (each prefix cut \emph{strictly before} its
target), mine ground truth \emph{within}/\emph{across} datasets plus embedding-similar candidates, and validate
by multi-pass re-checking and a five-annotator audit --- yielding graded, \emph{cross-instance} ground truth
with an oracle ceiling.}
\label{fig:pipeline}
\end{figure*}

\textbf{Construction} (Fig.~\ref{fig:pipeline}). From four source domains we run a vision-language model
(Gemini-2.5-Flash) to inventory objects and their timed state trajectories, generate the five question types, and mine
ground-truth instances both
\emph{within} and \emph{across} datasets (plus embedding-similar candidates). Candidates pass multi-pass
verification (independent multi-frame re-verification and object/state re-checks), and we additionally
release a high-precision \emph{consensus} pool (intersection of mining and verification). Prefixes are real
early clips so that the target state is strictly in the future. Full details and prompts are in the
appendix; all scripts are released.

\textbf{Human validation.} Because the GT is machine-mined, we validate it with humans. \textbf{Five}
annotators (in-house graduate researchers; \S Ethics) independently judged a stratified sample spanning all
difficulty tiers and question types --- both mined GT positives and top-ranked retrieved candidates ---
marking each as correct / incorrect / uncertain depiction of the queried future state (``uncertain''
counted as incorrect, the conservative choice). We report, on two distinct sub-populations of the $500$-query
sample, \emph{GT precision} (fraction of the \emph{mined-GT positives} confirmed correct) and a
\emph{suspected-miss rate} (fraction of \emph{retrieved non-GT} items judged correct, i.e.\ recall gaps).
GT precision is \textbf{$89.4\%$} (approx.\ $95\%$ Wilson CI $[86,92]$), the suspected-miss rate is
\textbf{$4.8\%$}, and agreement is substantial (Fleiss $\kappa=0.76$ over the 3-way label; note $\kappa$ is
somewhat optimistic under the ``correct''-heavy class balance). The pools are
nonetheless reliable with few corrections; the consensus pool (the intersection of mining and verification,
as above) is released for the highest-precision setting. Because that $4.8\%$ suspected-miss rate means
mining occasionally \emph{omits} a correct instance, every reported Recall@$k$ is a conservative \emph{lower
bound} (missed positives can only depress recall), so the method--ceiling gap we analyse is if anything
under-stated.

\textbf{No temporal leakage and quality.} Prefixes are cut strictly \emph{before} the event: across
\emph{all} benchmark queries the target state starts after the prefix ends in \textbf{100\%} of cases (a
construction guarantee; median prefix-to-target gap $1.9$\,s). Consistently, prefix-only baselines that do
not predict (persistence, zero-shot encoders) score far below the oracle (R@5 $\le 11$ vs.\ $53$): the prefix
does not reveal the answer, and the task genuinely requires prediction. Because states, questions, and GT are VLM-assisted, the benchmark can inherit generator biases;
we mitigate this with two independent verification passes, a consensus pool, and the human study above, and
flag it as a limitation.

\subsection{Dataset Composition and Statistics}
\textbf{Difficulty tiers.} The corpus mixes instances from all sources, so for each query the GT is split by
\emph{where the correct instance lives}: \emph{in-video} (the query's own video, easiest), \emph{in-dataset}
(a different video, same domain), and \emph{cross-dataset} (a different domain, the honest generalization
test); Fig.~\ref{fig:gtpool} shows an example query with its three-tier pool. We further provide
subject+predicate (type1) vs.\ predicate-only (type2) targets and the consensus pool, enabling robustness
checks to GT definition.

\textbf{Statistics.} The benchmark has 20{,}182 queries drawn from $5{,}061$ prefixes (one per
(video, object) pair) over 1{,}448 source video clips (train/val/test $14{,}390/2{,}996/2{,}796$,
sample\_id-disjoint to prevent video leakage). The primary retrieval corpus is \emph{video}: exactly
$218{,}276$ video segments ($51{,}362$ unique predicate states), each segment represented for retrieval by
its middle frame (visual space) or canonical state text (semantic space). Because a future \emph{end}-state
can be depicted across modalities (\emph{shattered} as a video clip or a still image), we also ship an
\emph{optional} $333$k-image corpus (VG $+$ MIT-States) as a cross-modal test --- which turns out to be hard
precisely for the \emph{dynamic} states (\emph{falling}, \emph{being pushed}) that only video fully captures
(App.~\ref{app:image}). The full release is $\approx$$28$\,GB: the $20{,}182$ query/GT annotations
($45$\,MB), the $5{,}061$ cut prefix clips ($18$\,GB), and the retrieval corpora ($\approx$$10$\,GB total
--- a $5.0$\,GB video corpus of $218{,}276$ segments and a $5.1$\,GB image corpus of $332{,}707$ images,
each shipped with precomputed feature indices). Horizons run $1$--$7$, near-skewed ($56\%$ $h{=}1$); we fold
the tiny $h{\ge}6$ tail into $h{=}5$ and report $h1$--$h5$. The corpus draws on four \emph{source datasets}
(\textsc{howtochange}~\citep{xue2024howtochange}/\textsc{oops}~\citep{epstein2020oops}/\textsc{ssv2}~\citep{goyal2017something}/\textsc{most};
Fig.~\ref{fig:stats}(c)) that define the difficulty tiers, spanning everyday themes (cooking, household,
sports, driving, animals). A full statistics table is in the appendix.

\section{LFTR: A Latent-Rollout Retriever}
\label{sec:method}

\begin{figure*}[t]\centering
\includegraphics[width=\textwidth]{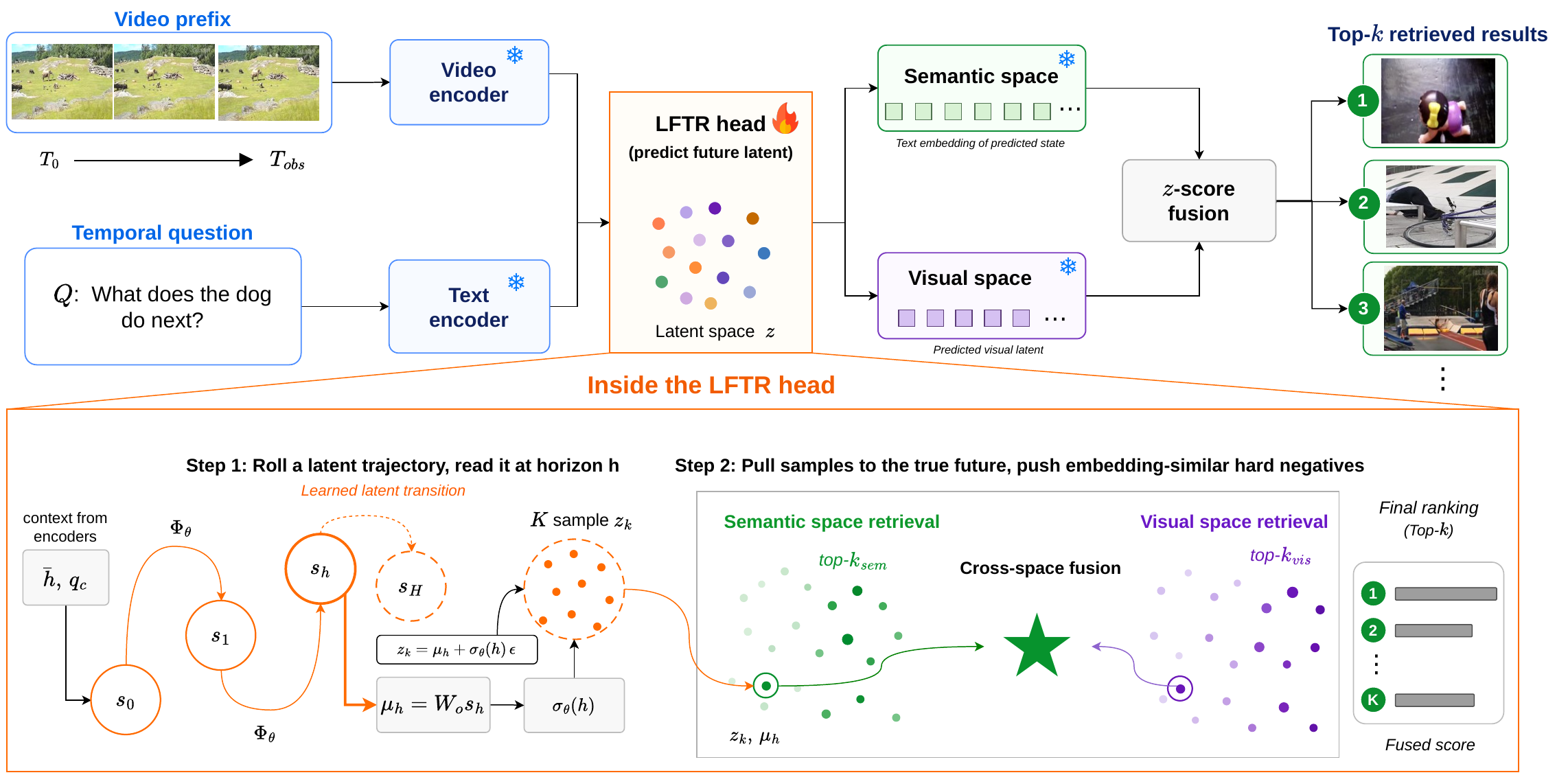}
\caption{\textbf{The LFTR method.} \emph{(Top)} Frozen video and text encoders feed the \emph{only} trained
module ($\sim$12M), the \textbf{LFTR head}, which predicts the future latent and reads it into a
\textbf{semantic} and a \textbf{visual} space; the two scores are \textbf{$z$-score fused} to return the
top-$k$ \emph{cross-instance} results. \emph{(Bottom) Inside the head.} \textbf{Step 1:} roll a latent
trajectory to horizon $h$ and sample $z_k{=}\mu_h{+}\sigma_\theta(h)\epsilon$. \textbf{Step 2:} retrieve by
max-similarity in each space, then fuse. Encoders stay frozen throughout.}
\label{fig:method}
\end{figure*}

\subsection{Model and inference}
LFTR rolls a latent trajectory of the future and reads it at the question's horizon (Fig.~\ref{fig:method}).
A \emph{frozen} encoder $E$ maps the 8-frame prefix to tokens; a small Transformer forms a context $\bar h$
that initializes $s_0$, and the question is embedded as $q_c = W_q\,\mathrm{BGE}(\text{question})$. A
transition operator rolls the latent forward,
\begin{equation}
s_{j+1} = s_j + \Phi_\theta(s_j,\ \bar h,\ q_c), \qquad j=0,\dots,H{-}1,
\end{equation}
and we read at the query horizon, $\mu_h = W_o\, s_h$. To model growing predictive uncertainty we
emit $K$ samples $z_k = \mu_h + \sigma_\theta(h)\,\epsilon_k$ and score a corpus item $m$ by multi-query
max-similarity $\max_k \cos(z_k, e_m)$. We keep the multi-step rollout for generality but ablate it
(Sec.~\ref{sec:analysis}) to be \emph{non-essential}: a single horizon-conditioned read ($H{=}1$) performs
comparably.

\textbf{Cross-space fusion.} An object's future state can be matched both \emph{semantically} (a text-state
space, BGE) and \emph{visually} (a DINOv2 image space). We run LFTR in both and z-score--fuse the per-item
scores; this complementary fusion is our strongest configuration.

\textbf{Naming.} ``LFTR'' denotes the \emph{recipe} --- a hard-negative-trained semantic-state head, read into
a retrieval space and fused across a semantic and a visual space --- which is what produces the gain; the
rollout (the ``R'') is kept only because one forward pass yields the whole future trajectory, useful for the
trajectory-richness analysis (App.~\ref{app:rolldistinct}), not because it improves accuracy. LFTR is thus a
strong, honestly-analyzed baseline built from known components (hard-negative InfoNCE, z-score fusion), not a
novel predictor.

\subsection{Training}
\textbf{What is a training sample.} The query side is the short prefix \emph{video} (8 frames $\rightarrow$
8 tokens). Each \emph{corpus/candidate} item---positive or negative---is a completed future \emph{segment}
represented by its \emph{middle frame} (a single still image) in the visual space, or by its canonical
state text in the semantic space; positives and negatives are thus single frames (or state strings)
standing for future segments, not full clips. Hard negatives are mined from the entire
$218$k-segment \emph{cross-instance} corpus as the nearest neighbours of the positive in the target space,
\emph{excluding the query's ground-truth set}. They are therefore other instances---overwhelmingly
\emph{different videos} (in our audited samples $47/48$), occasionally a different-state segment of the
same source clip---never the query's own labelled future.

\textbf{Objective.} Only the lightweight head ($\sim$12M parameters) is trained, with a multi-query InfoNCE
objective using random and hard (embedding-similar) negatives plus a grounding term that anchors $\mu_h$ to
the ground-truth embedding; the encoder remains frozen and its features are precomputed.

\section{Experiments}
\label{sec:exp}

\subsection{Experimental setup}
\textbf{Protocol.} We train on the train split, select on validation R@5, and report on the fair test set of
$n{=}2{,}430$ queries whose ground truth is present in the retrieval space (of $2{,}796$ total test queries;
the $366$ dropped have no GT segment in the target corpus). \emph{All main results retrieve over the video
corpus} (the primary benchmark); the image corpus is an optional cross-modal probe (App.~\ref{app:image}). The
encoder is frozen.
\textbf{Reporting convention.} A value with ``$\pm$'' is a $5$-seed mean$\pm$std; a bare value is a seed-0
single run (per-table protocol and a seed-0-vs-mean crosswalk are in the Table~\ref{tab:full} caption).
The baselines are (i) naive (random, question-only, kNN-future, persistence), (ii) zero-shot frozen encoders,
(iii) trained heads and LoRA fine-tuned encoders, and (iv) \textbf{M3}, our \emph{zero-shot} 32B multimodal-LLM
baseline. LFTR is instead a \emph{trained} retriever ($\sim$$12$M params), so the comparison is a cheap trained
retriever vs.\ a zero-shot 32B MLLM (M3) --- a \emph{different} family from the Gemini-2.5-Flash
generator, so unconfounded by same-family affinity. \textbf{Oracles (glossary).} All encode the \emph{true} future state and retrieve. We use three: the
\emph{semantic-text} oracle (state text $\rightarrow$ BGE, R@5 $53$), our clean ceiling; the
\emph{state-prototype} oracle (mean DINOv2 frame, R@5 $59$), \emph{mildly optimistic}; and the
\emph{independent-appearance} oracle (observed future frame, $\le6.6$), probing raw pixels. ``Oracle ceiling''
means the semantic-text $53$.

\subsection{Main results}

\begin{table*}[t]\centering\small\renewcommand{\arraystretch}{0.9}
\caption{Main results (test $n{=}2{,}430$). Ours in bold; LFTR is a five-seed mean. R@$k$ is \emph{micro};
``macro'' is R@5 averaged \emph{unweighted} over horizons $1$--$5$. ``---'': macro not computed for the naive
baselines.}
\label{tab:main}
\begin{tabular}{lccccc}
\toprule
Method & R@1 & R@5 & R@10 & MRR & macro \\
\midrule
Random & 0.0 & 0.0 & 0.0 & 0.0 & 0.0 \\
kNN-future & 1.5 & 4.5 & 5.6 & 3.0 & --- \\
Question-only & 1.6 & 5.5 & 8.5 & 3.8 & --- \\
Zero-shot (Qwen-VL, frozen) & 2.6 & 7.4 & 11.1 & 5.2 & --- \\
Text-only predictor (32B) & 3.9 & 9.3 & 12.5 & 6.6 & --- \\
Persistence & 3.9 & 10.8 & 14.0 & 7.4 & --- \\
M3 (Qwen3-VL-32B) & 5.2 & 11.9 & 15.4 & 8.6 & 7.7 \\
\midrule
LFTR $\rightarrow$ BGE (ours) & 4.1 & 11.3 & 15.0 & 7.6 & 7.5 \\
LFTR $\rightarrow$ EmbeddingGemma (ours) & 4.7 & 12.3 & 16.3 & 8.4 & 8.4 \\
LFTR $\rightarrow$ SigLIP (ours) & 4.5 & 10.6 & 14.1 & 7.6 & 7.7 \\
LFTR $\rightarrow$ LanguageBind (ours) & 4.6 & 11.0 & 14.5 & 7.8 & 7.0 \\
LFTR $\rightarrow$ InternVideo2 (ours) & 4.4 & 11.4 & 15.2 & 8.0 & 7.9 \\
LFTR $\rightarrow$ DINOv2 (ours) & 5.0 & 12.8 & 16.4 & 8.8 & 8.3 \\
LFTR $\rightarrow$ LingBot-Vision (ours) & 5.8 & 13.6 & 17.3 & 9.6 & 9.2 \\
\textbf{LFTR-fusion (BGE$+$DINOv2, ours)} & \textbf{6.8} & \textbf{16.1} & \textbf{20.3} & \textbf{11.4} & \textbf{10.8} \\
\midrule
Oracle-BGE (ceiling) & 23.9 & 53.0 & 67.8 & 37.6 & 51.5 \\
Oracle-DINOv2 (ceiling) & 48.3 & 59.1 & 63.2 & 53.5 & 59.8 \\
\bottomrule
\end{tabular}
\end{table*}

Table~\ref{tab:main} shows LFTR-fusion reaches $16.1{\pm}1.0$ R@5, a \emph{significant} improvement over
the zero-shot 32B MLLM (paired bootstrap $+4.81$pp, $95\%$ CI $[3.00,6.58]$, $p<10^{-4}$), at $\sim$$12$M
trainable parameters and $\sim$$0.06$\,ms/query --- four orders of magnitude cheaper than a $32$B generator
(App.~\ref{app:eff}). Stacking further \emph{complementary} spaces reaches R@5 $18.6{\pm}0.6$
(App.~\ref{app:fusion}); we keep the $2$-space fusion as the headline. The naive \emph{persistence} baseline is
surprisingly strong ($10.8$): a method must \emph{predict}, not copy the present, to win. By question type,
fusion wins on every one and
the far-horizon end-state (Q2) is hardest for all learned methods while the Oracle stays flat
($\sim$$51$--$56$; Table~\ref{tab:qtype}, App.~\ref{app:qtypes}), locating the difficulty in \emph{far-horizon
prediction} rather than a question form.

\textbf{Is the comparison to M3 fair?} We stress-test fairness two ways (Table~\ref{tab:m3fair}), and neither
closes the gap. \emph{(a)} Giving M3 the \emph{strongest} text space (EmbeddingGemma) does \emph{not} help it
($11.2\approx11.4$) though it helps LFTR ($11.3{\to}12.3$): a short predicted \emph{string} cannot exploit a
richer semantic space, a learned latent can. \emph{(b)} Giving M3 a CLIP text$\rightarrow$image path, or
fusing, is weak or \emph{hurts} ($3.1$--$10.3$), because a predicted string is a poor visual query. LFTR-fusion
wins by predicting a \emph{native visual latent} no text-predicting MLLM can reach with \emph{any} text
encoder --- the advantage is methodological, not ``more spaces''.

\begin{table}[t]\centering\small
\caption{Fair MLLM comparison (test R@5). Giving M3 the \emph{best} text space, a visual path (CLIP
text$\rightarrow$image), or fusing does not close the gap; predicting a native visual latent (LFTR) does. Rows
use the \emph{intersection} subset, so M3$\rightarrow$BGE is $11.4$ here vs.\ $11.9$ in Table~\ref{tab:main}.}
\label{tab:m3fair}
\begin{tabular}{lc}
\toprule
Method & R@5 \\
\midrule
M3 $\rightarrow$ BGE (text-state) & 11.4 \\
M3 $\rightarrow$ EmbeddingGemma (best text) & 11.2 \\
M3 $\rightarrow$ CLIP (text$\rightarrow$image) & 3.1 \\
M3 fusion (BGE $+$ CLIP) & 10.3 \\
M3 fusion (Gemma $+$ CLIP) & 9.9 \\
\midrule
\textbf{LFTR-fusion (BGE $+$ DINOv2, ours)} & \textbf{16.1} \\
\bottomrule
\end{tabular}
\end{table}

\subsection{Discussion}
\label{sec:analysis}
\textbf{Where the gain comes from (honest ablation).} Three effects are robust across five seeds.
(i)~\emph{Cross-space fusion} is the largest: fusing the semantic (BGE) and visual (DINOv2) spaces lifts R@5
from $\sim$$12.8$ to $16.1$, and what matters is \emph{complementarity, not count} --- a redundant same-type
space adds nothing, whereas stacking complementary strong spaces (BGE, Gemma, DINOv2,
boundary-SSL LingBot) reaches $18.6{\pm}0.6$, a significant $+2.5$pp over the 2-space fusion
(App.~\ref{app:fusion}). (ii)~\emph{Hard negatives} give a large, monotone gain \emph{in the visual space} ---
mining embedding-similar (visually confusable but wrong) future states adds $+3.9$pp R@5 ($p<10^{-3}$),
saturating near $128$ (App.~\ref{app:hardneg}) --- but are \emph{neutral} in the text-state space, where states
are already lexically separated: visual, not semantic, confusability is the binding constraint.
(iii)~Question conditioning, grounding, the uncertainty head, and $K/H$ each move a single-space model within
seed noise ($\pm0.7$).

\textbf{The multi-step rollout is not the source of the gain (honest negative).} Under identical training the
full $H{=}8$ trajectory and an $H{=}1$ horizon-conditioned read are statistically indistinguishable
($12.80{\pm}0.72$ vs.\ $12.40{\pm}1.02$, $p{=}0.54$; n.s.\ at every horizon), and intermediate-horizon
supervision does not help. The gain comes from the future-state head, hard-negative visual training, and
fusion --- not the rollout mechanism.

\textbf{The rollout is nonetheless a rich, under-exploited trajectory.} It is not inert: one pass emits the
whole future trajectory, and an oracle reading the \emph{best-per-query step} reveals a large
$+4.3$/$+4.8$pp of untapped signal in both spaces --- turning that into a robust, oracle-free read rule is an
open problem the benchmark exposes (App.~\ref{app:rolldistinct}).

\textbf{Encoders and tiers.} LFTR is encoder-agnostic across eight frozen encoders, and a strong visual
target can beat the text-state space (App.~\ref{app:matrix}). By difficulty tier, in-video retrieval is solved
by appearance \emph{persistence} (R@5 $75$ vs.\ $\le5$ for all semantic methods, \emph{including} the Oracle)
--- exactly why we score \emph{cross-instance}; the cross-dataset tier ($\approx8.4$) is the honest
generalization gap.

\textbf{The task is predictor-bound: forecasting, not perception, is the bottleneck.} Retrieval with the
\emph{true} future state reaches R@5 $53.0$ (BGE text) and $59.1$ (DINOv2 prototype), roughly flat across
horizon (Table~\ref{tab:ceiling}). Crucially, an \emph{independent} oracle that queries with the object's
\emph{actual observed future frame} and retrieves cross-instance is near the floor in \emph{every} visual
space (DINOv2 $5.7$, SigLIP $6.6$, CLIP $3.0$, LingBot $6.5$; App.~\ref{app:oraclevis}): the same state looks
very different across videos, so the task cannot be solved by raw appearance and demands a \emph{semantic}
abstraction. Given that abstraction the future is highly retrievable ($53$--$59$), yet the best learned predictor
realises only $\sim$$16$--$18$, and every method collapses beyond $h{=}2$. Two controls locate the residual in
\emph{forecasting}: a sighted $32$B MLLM (M3) reaches only $11.9$ ($\sim$$41$pp below the ceiling), and a
$7$B$\to$$32$B text-LLM scale ladder forecasts to only $6.5{\to}9.3$ --- so neither perception nor scale is
the missing ingredient. The
raw $16{\to}53$ gap is partly a \emph{retrieval-canonicalization} channel (free-form vs.\ canonical
string); the genuine \emph{forecasting} gap a better predictor can close is $\sim$$14$pp ($16{\to}30$;
App.~\ref{app:channel}), the core learnable open problem PSR poses.

\begin{table}[t]\centering\small
\caption{Ceiling decomposition: R@5 by horizon ($n{=}1356/576/312/109/49$ for $h{=}1{-}5$). The
\emph{semantic}-state oracles are high and roughly flat, but an \emph{independent} oracle using the observed
future \emph{frame} is near the floor: the task is tractable semantically, not by appearance. ``all'' is the
micro mean; far-horizon cells are noisy.}
\label{tab:ceiling}
\resizebox{\columnwidth}{!}{%
\begin{tabular}{lcccccc}
\toprule
Oracle & all & $h1$ & $h2$ & $h3$ & $h4$ & $h5$ \\
\midrule
Semantic: DINOv2 prototype & 59.1 & 59.8 & 61.5 & 51.0 & 61.5 & 65.3 \\
Semantic: BGE text (clean) & 53.0 & 52.9 & 53.3 & 54.5 & 45.9 & 51.0 \\
Visual: observed future (indep.) & 5.7 & 7.4 & 5.3 & 3.3 & 1.7 & 0.0 \\
\midrule
\emph{LFTR} & \emph{12.8} & \emph{14.7} & \emph{14.2} & \emph{6.0} & \emph{3.9} & \emph{2.9} \\
\bottomrule
\end{tabular}}
\end{table}

\textbf{Multi-object and qualitative behaviour.} LFTR transfers to the multi-object setting per object, though
\emph{joint} all-objects-correct is combinatorial and near zero even for the Oracle, an open regime
(App.~\ref{app:q5}). Qualitatively, fusion recovers the correct future where the naive and MLLM baselines
return a visually similar but \emph{wrong} (hard-negative) state, and failures are almost always a
\emph{semantically adjacent} state (\emph{rolling on floor}$\rightarrow$``falling''; \emph{lying on
ground}$\rightarrow$``being pushed'') --- precisely the fine-grained discrimination hard-negative training
targets \citep{robinson2021hardneg} (examples in App.~\ref{app:qual},~\ref{app:hardneg}).

\section{Conclusion}
\label{sec:concl}
We pose Predictive State Retrieval --- predict an object's future state from a prefix and retrieve
cross-instance examples across horizons --- and show that a cheap frozen-encoder retriever, whose gain comes
from cross-space fusion and hard negatives (\emph{not} the latent rollout, which we find null), beats a
\emph{zero-shot} 32B MLLM at a fraction of the cost. Its ceiling decomposition isolates
\emph{far-horizon forecasting} as the core learnable bottleneck --- one of several measurable open problems
(cross-dataset, multi-object) with large headroom that make PSR a useful forward-looking benchmark.

PSR's oracle ceiling and cheap baseline make it a testbed for the two open problems it exposes: an
oracle-free read of the \emph{under-exploited} rollout, and a \emph{cross-modal} bridge for \emph{dynamic}
states.

\clearpage
\section*{Limitations}
The image modality is hard even for the Oracle (R@5 $8.9$) due to a vocabulary mismatch between
video-derived queries and image-corpus states; joint multi-object and far-horizon prediction remain open.
Methodologically, our headline gain is driven by cross-space
fusion and hard-negative visual training rather than by a fundamentally new predictor: a controlled
isolation shows the multi-step latent rollout performs no better than a single horizon-conditioned read,
and intermediate-horizon supervision does not help. Even end-to-end LoRA fine-tuning of the prefix encoder
yields only a small, within-noise gain ($+0.4$--$0.5$ R@5 across DINOv2 and CLIP) at $\sim$$10^{3}\times$ the
training cost (App.~\ref{app:unfreeze}), so the frozen prefix is a \emph{mild} constraint, not the bottleneck.
We therefore present LFTR as a strong, cheap baseline and---given the flat, high Oracle---position closing
the far-horizon \emph{prediction} gap as the central future work.

We flag five validity threats explicitly. \textbf{(i) Single-generator authorship.} States, questions, and
GT candidates are produced by one VLM (Gemini-2.5-Flash). Two facts bound the resulting confound: our M3
baseline (Qwen3-VL-32B) is a \emph{different} family from the generator, so the M3 comparison is not
confounded by same-family affinity (cross-family phrasing could in principle penalize M3, but its $11.9$ sits
far below the $\sim$$30$ free-form ceiling, so it is forecasting-, not phrasing-limited; App.~\ref{app:channel});
and a stronger automated judge (Gemini-3-Flash) agrees with the human
labels at $93.4\%$ (Cohen's $\kappa{=}0.78$) and estimates GT precision at $88.6\%$, matching the human
$89.4\%$ (App.~\ref{app:crossmodel}). The residual concern is that a single generator's phrasing conventions
may still shape the GT; because that cross-model judge shares the generator's Gemini family, a fully
independent cross-\emph{family} re-authoring of a held-out slice remains future work. \textbf{(ii) The appearance floor is partly by construction.}
Because GT links instances by shared \emph{semantic} state, a visual-appearance oracle is scored against a
semantically-defined positive set; ``appearance cannot solve PSR'' should be read as ``cannot recover the
\emph{semantic-state} GT,'' not as a claim about a hypothetical visually-defined GT. \textbf{(iii) Ceiling
partition.} Both semantic oracles are, by our own audit, mildly optimistic, so while a \emph{large}
method--ceiling gap is robust, we do not claim an exact split of that gap into prediction vs.\ retrieval
error. \textbf{(iv) Human study scope.} Our validation is $500$ queries by five in-house annotators (complemented by
the automated cross-model check of App.~\ref{app:crossmodel}); it establishes GT precision but is not a
stratified, per-tier human audit, and there is no measured
\emph{human} performance on the retrieval task itself (so the $16{\to}53$ headroom is an oracle bound, not a
demonstrated human-achievable target). \textbf{(v) Generalization} is cross-\emph{source} transfer within
four domains, not a leave-one-domain-out generalization study.

\section*{Ethics Statement}
The benchmark is built on publicly released video datasets (\textsc{oops}, \textsc{ssv2}, \textsc{most},
\textsc{howtochange}) and is redistributed only as query/GT annotations and precomputed feature indices,
under the source datasets' licenses; we release no raw video we do not have the right to. States, questions,
and candidate GT are produced with a vision-language model and may contain errors or social biases; we
mitigate this with two independent verification passes, a high-precision consensus pool, and a five-annotator
human study (Sec.~\ref{sec:bench}), and we flag residual model bias as a limitation. The five annotators were
volunteer graduate researchers fluent in English who judged only object-state depictions (no personal or
sensitive attributes); they gave informed consent, were not compensated per-item (to avoid speed incentives),
and each labelled the same stratified $500$-query sample, from which we report Fleiss $\kappa=0.76$
(substantial, though optimistic under the ``correct''-heavy class balance; \S\ref{sec:bench}). The task
targets retrieval of everyday object states and is not intended for surveillance or person identification.

\section*{Reproducibility}
We release the dataset (queries, graded GT pools, splits), the 218k-segment corpus index, all baseline and
method code, training configurations, and the evaluation scripts; every number in the paper is regenerated
from released eval artifacts.

\bibliography{custom}

\appendix

\section{Benchmark and Protocol Details}
\label{app:bench}

\textbf{State extraction and canonicalization.} For each segment the VLM (Gemini-2.5-Flash) emits a free-form
future-state description, which we parse into a \emph{predicate-level} state (\emph{type2}, e.g.\
\emph{falling}) and a \emph{subject$+$predicate} state (\emph{type1}, e.g.\ \emph{ball falling}). Free-form
descriptions are lexically noisy --- \emph{falling}, \emph{tumbling down}, and \emph{dropping} denote the
same state --- so we \emph{canonicalize} synonyms by greedy BGE-embedding clustering: states are visited
\emph{most-frequent first}, and each is merged into an existing canonical cluster if its BGE cosine
similarity to the cluster seed is ${\ge}\,0.85$, otherwise it seeds a new canonical state (the threshold was
chosen on a pilot sweep over $\{0.80,0.85,0.90\}$ to balance merge rate against semantic coherence, keeping
mean cluster size ${\le}5$). Canonicalization reduces the raw state vocabulary to $51{,}362$ canonical
predicate-level states, and every corpus segment and query target is re-keyed to its canonical state; all
retrieval and ground truth are defined at this canonical level.

\textbf{Ground-truth mining.} For a query whose target is canonical state $s$, we mine three complementary GT
pools: \emph{in-dataset} (a different video of the \emph{same} domain whose segment carries canonical state
$s$), \emph{cross-dataset} (a segment from a \emph{different} domain with canonical state $s$ --- the honest
cross-instance test), and \emph{bge-similar} (BGE neighbours of $s$ at cosine ${\ge}\,0.85$, a looser
recall-oriented pool). Candidates then pass multi-pass VLM re-verification (independent multi-frame object /
state re-checks); the released high-precision \emph{consensus} pool is the intersection of the mining and
verification passes. Structured hard negatives for training are the nearest BGE / visual neighbours of the
target that are \emph{excluded} from every GT pool (App.~\ref{app:hardneg}). Because mining is embedding-based
and canonical, a $4.8\%$ suspected-miss rate remains (\S\ref{sec:bench}), so every Recall@$k$ is a
conservative lower bound.

\textbf{Single-prefix protocol.} Each (video, object) contributes \emph{one} fixed early prefix (8 frames
before the event); multiple queries---across question types and horizons---share that prefix, so a method
must \emph{predict} rather than exploit a late, near-answer prefix. (We adopt this over the alternative of
cutting a fresh near-answer prefix per horizon, which would leak the target; we use the single-prefix
protocol throughout.) Splits: train $14{,}390$ / val $2{,}996$ /
test $2{,}796$; the fair test set with ground truth present in the retrieval space is $n{=}2{,}430$.
Question types are Q1 (next state), Q2 (end state), Q3 (after $k$ seconds), Q4 (intermediate state), plus
Q5 (multi-object end states, App.~\ref{app:q5}). The corpus holds $218{,}276$ \emph{video segments} (short
watchable clips, each represented for retrieval by its middle frame) over $51{,}362$ unique predicate-level
states from four source domains (\textsc{howtochange}, \textsc{oops}, \textsc{ssv2}, \textsc{most}). An
optional image corpus (Visual Genome~\citep{krishna2017vg} $+$ MIT-States~\citep{isola2015states}) is also
provided (App.~\ref{app:image}).

\textbf{Difficulty tiers (where the correct future instance lives).} For each query the ground-truth pool
is partitioned into: \emph{in-video (self)}---the query's own source video's future segment (future
$\approx$ present \emph{visually}); \emph{in-dataset}---a different video, same domain; and
\emph{cross-dataset}---a different domain, which forces matching the abstract state and is the honest
generalization test. The main benchmark \emph{excludes} the self tier from the retrieval pool to force
cross-instance prediction (App.~\ref{app:tiers}).

\textbf{Ground-truth definitions.} States come in two granularities: \emph{type2} (predicate only, the
default index) and \emph{type1} (subject $+$ predicate, stricter). A high-precision \emph{consensus} pool
(the intersection of mining and verification, \S\ref{sec:bench}) is also provided. Method rankings are stable across all definitions
(App.~\ref{app:gtrobust}).

\begin{figure*}[t]\centering
\includegraphics[width=\textwidth]{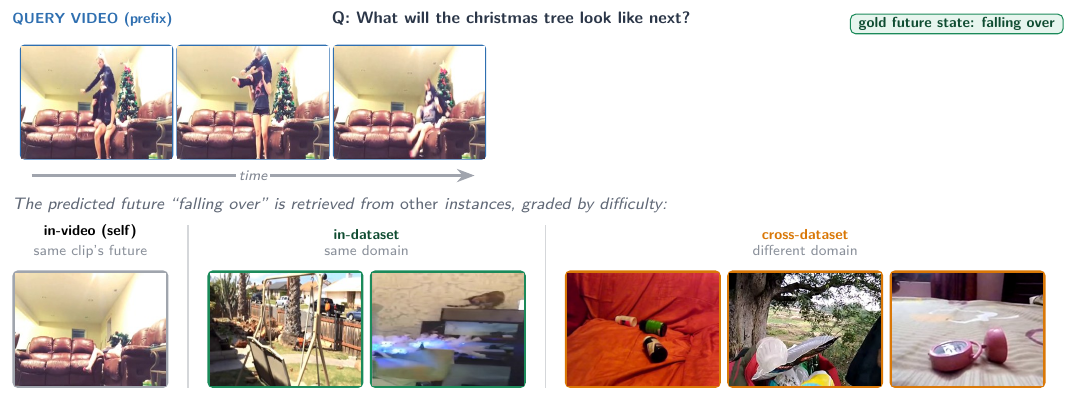}
\caption{\textbf{Graded ground truth.} The observed prefix (blue, ordered in time) and its predicted future
state (\emph{falling over}) as it appears in \emph{other} instances, graded by difficulty: the query's own
video (\textbf{in-video}, solved by appearance), another video of the same domain
(\textbf{\textcolor{green!55!black}{in-dataset}}), and a different domain
(\textbf{\textcolor{orange!80!black}{cross-dataset}}). The main benchmark scores cross-instance retrieval
and excludes the in-video tier.}
\label{fig:gtpool}
\end{figure*}

\section{Cross-Model GT Validation}
\label{app:crossmodel}
Beyond the five-annotator human study (\S\ref{sec:bench}), we cross-check the same $500$-query sample with a
\emph{stronger, automated} judge --- Gemini-3-Flash --- prompted with the identical
correct\,/\,incorrect\,/\,uncertain rubric. It agrees with the human majority label at $93.4\%$
(Cohen's $\kappa{=}0.78$, on par with the $5$-annotator $\kappa{=}0.76$) and independently estimates GT
precision at $88.6\%$, corroborating the human $89.4\%$ (Table~\ref{tab:crossmodel}). This thickens the
reliability evidence with an automated second opinion. Because the judge shares the generator's Gemini
family, it does \emph{not} substitute for a fully independent cross-\emph{family} re-authoring
(Limitation~(i)); the M3 baseline (Qwen3-VL-32B), however, \emph{is} cross-family from the generator, so the
headline method-vs-MLLM comparison is already unconfounded by same-family affinity.

\begin{table}[t]\centering\small
\caption{\textbf{Cross-model GT validation.} Gemini-3-Flash re-scores the $500$-query human-validated sample:
its GT-precision estimate matches the human audit, and it agrees with the human 3-way labels at $93.4\%$
(Cohen's $\kappa{=}0.78$).}
\label{tab:crossmodel}
\begin{tabular}{lc}
\toprule
 & value \\
\midrule
Human GT precision (5 annotators) & $89.4\%$ \\
Gemini-3-Flash GT precision & $88.6\%$ \\
\midrule
Human--Gemini agreement (3-way label) & $93.4\%$ \\
Cohen's $\kappa$ (human vs.\ Gemini) & $0.78$ \\
\bottomrule
\end{tabular}
\end{table}

\section{Question-Type Taxonomy}
\label{app:qtypes}
\begin{figure*}[t]\centering
\includegraphics[width=\textwidth]{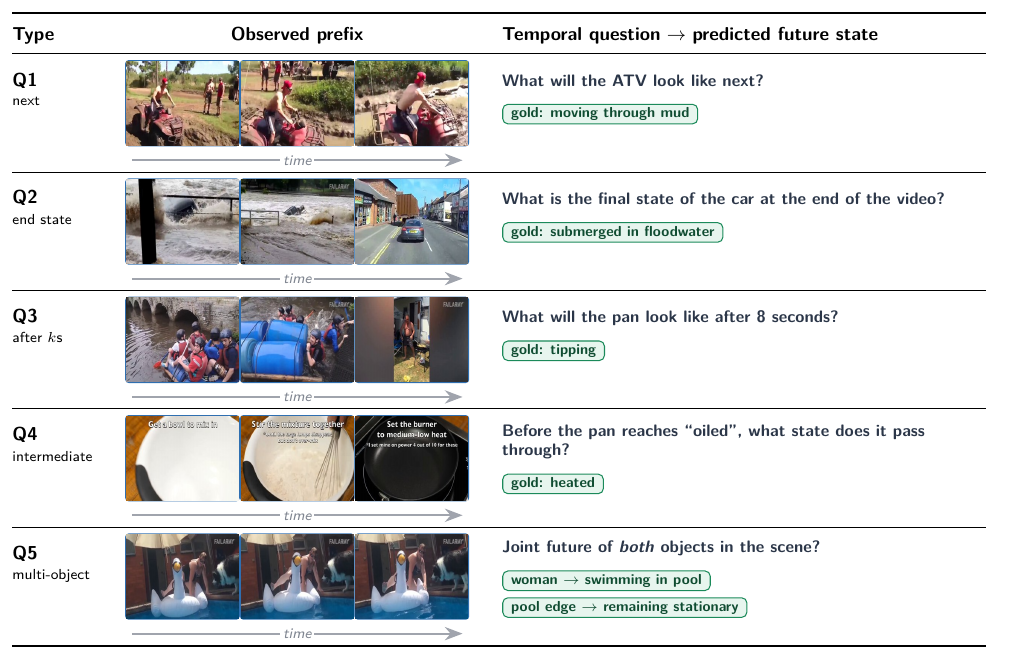}
\caption{\textbf{One worked example per question type} (Q1--Q5). Each row shows the observed prefix (blue
frames, ordered left-to-right in time), the temporal question, and its \emph{predicted} future state (gold).
The types probe different horizons: \textbf{Q1} the immediate next state, \textbf{Q2} the end state,
\textbf{Q3} the state after $k$ seconds, \textbf{Q4} an intermediate state before a named later state, and
\textbf{Q5} the joint future of \emph{two} objects in one scene (multi-object, scored per object and jointly;
App.~\ref{app:q5}). The target never appears in the prefix; it must be predicted, then matched by
\emph{meaning} in other videos.}
\label{fig:qtype_examples}
\end{figure*}

PSR poses \emph{five} temporal question types over each (video, object) prefix, each probing a different
facet of future-state prediction (Table~\ref{tab:qtypes}; a worked example per type in
Fig.~\ref{fig:qtype_examples}). Q1--Q4 constitute the main benchmark
($20{,}182$ queries; one shared early prefix per (video, object), queried at several horizons and types);
Q5 is a separate multi-object stress test (App.~\ref{app:q5}). All types are generated by prompting a VLM
over each object's \emph{timed state trajectory} $\{(\text{state}_i, t_i)\}$ and are answered against the
same graded, cross-instance ground truth. Per-type retrieval results are in Table~\ref{tab:qtype}.

\noindent\textbf{Q1 --- Next state.} The \emph{immediately following} state after the prefix. Almost always
a single step ahead ($h{=}1$ for $98\%$ of Q1), so it is the shortest-horizon, most perceptually anchored
type, yet still requires predicting a \emph{transition} rather than copying the present.

\noindent\textbf{Q2 --- End state.} The object's \emph{final} state at the end of the video. Because ``the
end'' is a variable distance away, Q2 spans the \emph{full horizon range} ($h{=}1$--$7$) and is the
\emph{hardest} type (Table~\ref{tab:qtype}): the far future is genuinely multi-modal.

\begin{figure*}[t]\centering
\includegraphics[width=\textwidth]{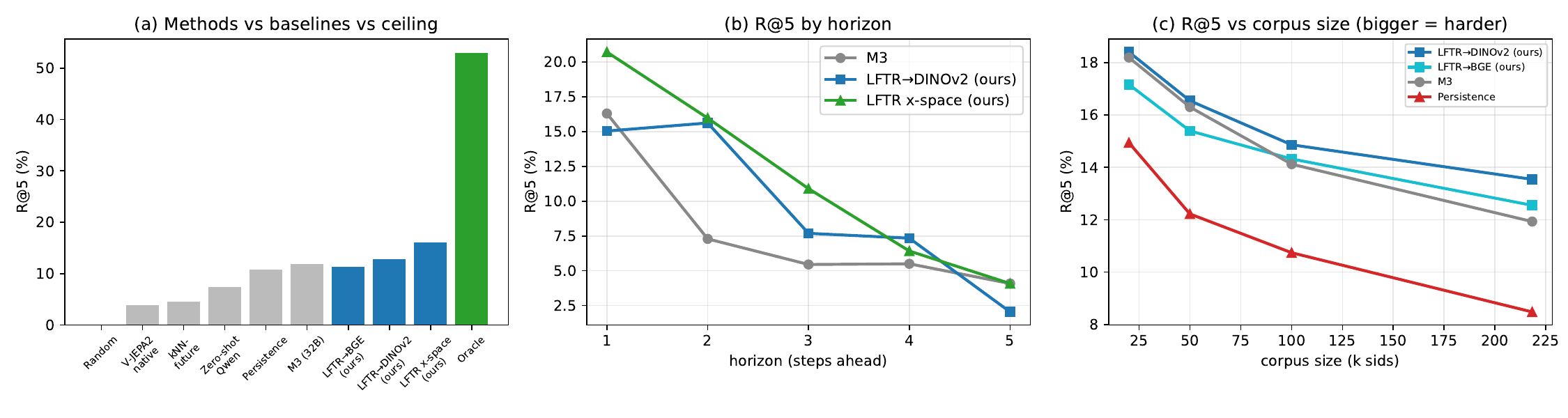}
\caption{(a) Methods vs.\ baselines vs.\ ceiling (5-seed means, matching Table~\ref{tab:main}). (b) R@5 by
horizon: LFTR holds the mid-horizon range where the MLLM collapses, but every learned predictor falls far
below the (horizon-flat) Oracle beyond $h{=}2$. (c) R@5 vs.\ corpus size: LFTR degrades more gracefully than
M3 as the database grows. Panels (b,c) show a representative single run, so absolute values run ${\sim}1$pp
above the seed-averaged Table~\ref{tab:main}.}
\label{fig:summary}
\end{figure*}

\noindent\textbf{Q3 --- After $k$ seconds.} The state a \emph{fixed} $k$ seconds ahead; $k$ maps to a
prediction horizon. Tests fixed-lead-time forecasting and, with $k$ varied, gives the horizon-stratified
curves in Fig.~\ref{fig:summary}(b).

\noindent\textbf{Q4 --- Intermediate state.} A state the object \emph{passes through before} reaching a
named later state (``before losing balance, what state does it pass through?''). The \emph{largest} type
($7{,}981$), it probes trajectory/ordering reasoning rather than endpoint prediction.

\noindent\textbf{Q5 --- Multi-object.} The \emph{joint} future states of $\ge2$ objects in the scene,
scored per object and jointly (App.~\ref{app:q5}); a compositional stress test kept separate from the
single-object main benchmark.

\begin{table*}[t]\centering\small
\caption{Question-type taxonomy and statistics. Q1--Q4 form the main benchmark; Q5 is a separate
multi-object evaluation (App.~\ref{app:q5}). ``$N$'' is train/val/test query counts; ``$h$'' is the
prediction-horizon coverage (steps ahead). Per-type R@5 is reported in Table~\ref{tab:qtype}.}
\label{tab:qtypes}
\resizebox{\textwidth}{!}{%
\begin{tabular}{lllcr}
\toprule
Type & Asks for & Example question $\to$ target & $N$ (tr/val/test) & $h$ \\
\midrule
Q1 Next state & the immediately next state & \emph{``What will the man look like next?''} $\to$ struggling with jellyfish & 3028/634/588 & $1$ ($98\%$) \\
Q2 End state & the final state at video end & \emph{``What is the final state at the end?''} $\to$ falling into water & 2883/582/555 & $1$--$7$ \\
Q3 After-$k$ & the state $k$ seconds ahead & \emph{``\dots\ look like after 3 seconds?''} $\to$ swinging object toward grill & 2784/585/562 & $1$--$7$ \\
Q4 Intermediate & a state passed through first & \emph{``Before losing balance, what state does it pass through?''} $\to$ struggling with jellyfish & 5695/1195/1091 & $1$--$7$ \\
\midrule
Q5 Multi-object & joint future of $\ge2$ objects & per-object end states (App.~\ref{app:q5}) & separate eval & --- \\
\bottomrule
\end{tabular}}
\end{table*}


\begin{table}[t]\centering\small
\caption{R@5 by question type (test). Fusion wins on every type; the end-state Q2 is hardest for all learned
methods while the Oracle is flat ($\sim$$51$--$56$), locating the difficulty in \emph{far-horizon}
prediction rather than a particular question form.}
\label{tab:qtype}
\resizebox{\columnwidth}{!}{%
\begin{tabular}{lcccc}
\toprule
Method & Q1 next & Q2 end & Q3 after-$k$ & Q4 inter. \\
\midrule
M3 (Qwen3-VL-32B) & 15.0 & 7.2 & 11.2 & 13.0 \\
LFTR $\rightarrow$ DINOv2 & 13.8 & 7.6 & 16.3 & 13.1 \\
\textbf{LFTR fusion (ours)} & \textbf{17.8} & \textbf{9.9} & \textbf{20.0} & \textbf{16.2} \\
\midrule
Oracle-BGE (ceiling) & 51.0 & 51.1 & 55.7 & 53.6 \\
\bottomrule
\end{tabular}}
\end{table}

\section{Implementation and Baseline Details}
\label{app:impl}
\textbf{Encoders (exact models).} Prefix/query encoders: \emph{Qwen-VL} (the vision tower of
Qwen3-VL-32B), \texttt{facebook/dinov2-large}~\citep{dinov2},
\texttt{openai/clip-vit-base-patch32}~\citep{radford2021clip}, and
\texttt{facebook/vjepa2-vitl-fpc64-256}~\citep{vjepa2}. Text is encoded with
\texttt{BAAI/bge-small-en-v1.5}~\citep{bge} ($384$-d).
The prefix is $8$ frames sampled from the early clip; each image encoder yields one token per frame
($8$ tokens). Every corpus segment is embedded for retrieval by its \emph{middle} frame in each visual
space; the semantic space embeds the $51{,}362$ canonical state strings with BGE. All encoder features are
precomputed and frozen.

\textbf{LFTR (full configuration).} Internal width $D_m{=}512$; a $3$-layer, $8$-head Transformer encoder
forms the context; the transition $\Phi$ is a $3$-layer MLP (depth $=3$); rollout depth $H{=}8$;
$K{=}32$ samples; a horizon-conditioned $\sigma$ head. Objective: multi-query max-similarity InfoNCE
($\tau{=}0.04$) with $384$ random $+$ $128$ hard negatives (nearest neighbours of the positive in the
target space, excluding the query GT), plus a grounding MSE (weight $0.5$) to the target embedding. AdamW
(lr $3{\times}10^{-4}$, weight decay $10^{-4}$), cosine schedule, $40$--$60$ epochs, batch $256$, best-val-R@5
selection, five seeds. Only the head ($\sim$$11.98$M params) is trained. Cross-space fusion runs LFTR in the
BGE and DINOv2 spaces and sums per-item $z$-scored scores.

\textbf{Baselines (exact setup).}
\begin{itemize}\itemsep1pt
\item \textbf{Random}: a random ranking of the corpus.
\item \textbf{Question-only}: BGE-encode the question text and retrieve in the BGE state space (no visual
input) --- isolates language priors.
\item \textbf{Text-only predictor}: a text LLM (Qwen2.5-VL-7B, run text-only) is given a symbolic transcript
of the observed prefix states (each state with its timestamp, truncated at the query time so the answer is
never revealed) plus the question, and outputs a short future-state phrase; we BGE-encode it and retrieve.
This isolates \emph{prediction} from perception: even with an oracle textual description of the past, the
future state is hard to forecast (R@5 $6.5$, only marginally above the question-only prior).
\item \textbf{kNN-future} (non-parametric): take the nearest training query by prefix features and use its
ground-truth future state as the prediction; no learned parameters.
\item \textbf{Persistence} (``future$=$present''): the \emph{last} observed prefix frame's embedding
(Qwen-VL or DINOv2) retrieves in the same space --- copies the present.
\item \textbf{Zero-shot}: mean-pool the $8$ prefix tokens of a frozen encoder and retrieve in that
encoder's corpus; no training, no prediction.
\item \textbf{Trained-head}: the LFTR head with $H{=}1$ (a horizon-conditioned read, no multi-step
rollout), trained identically; and LoRA-fine-tuned encoders.
\item \textbf{LoRA fine-tune}: LoRA ($r{=}16$, $\alpha{=}32$, targets $q/k/v$ ($+\text{out}$ for CLIP),
dropout $0.1$) on the encoder, mean-pool $\to$ projection $\to$ InfoNCE; batch $64$, $\sim$$15$ epochs.
\item \textbf{M3} (MLLM): Qwen3-VL-32B-Instruct is shown the $8$ prefix frames and the question and prompted
to name the object's future state in text; the prediction is BGE-encoded and retrieved in the BGE state
space. Fair visual variants (CLIP text$\rightarrow$image, and fusion) are in Table~\ref{tab:m3fair}.
\item \textbf{Oracle} (ceiling): encode the \emph{true} target state --- as text (BGE) or as the state's
mean DINOv2 frame-prototype (visual) --- and retrieve; the upper bound given perfect prediction.
\end{itemize}

\textbf{Scoring.} Cosine similarity; for LFTR the query score is the max over its $K$ samples; a state-space
method expands each retrieved state to its member segments. R@$k$/MRR follow the standard definitions on the
graded GT. \textbf{Compute.} The frozen-feature head trains in minutes on one GPU; end-to-end LoRA of the
encoder costs $\sim$$40$\,min/epoch (App.~\ref{app:unfreeze}).

\section{Full Baseline and Method Table}
\label{app:fulltable}
Table~\ref{tab:full} reports every family at all cutoffs on the fair test set ($n{=}2{,}430$). Highlights:
Qwen-VL is the strongest prefix encoder and a DINOv2 visual space the strongest retrieval target
(App.~\ref{app:matrix}); trained heads beat zero-shot and LoRA fine-tuning. Every other MLLM we tried ---
across \emph{families} and \emph{scales} --- trails the 32B Qwen-VL M3: an image MLLM (LLaVA-1.6-7B $4.2$), a
$27$B coding VLM ($4.6$), and a \emph{video}-native MLLM at two sizes (LLaVA-NeXT-Video $7$B$\to$$34$B:
$2.8\to4.2$). Even a $34$B video model reaching only $4.2$ --- \emph{below} persistence ($10.8$) ---
demonstrates PSR is not trivially solvable by a strong MLLM. Crucially, \emph{none} of these baselines ---
including the strongest, Qwen-VL M3 --- is the family that generated the GT (Gemini-2.5-Flash), so M3's lead
reflects genuine forecasting ability, not same-family affinity; the one residual question --- whether the
single Gemini generator's phrasing conventions advantage some forecasters --- would be settled by a full
cross-family re-authoring (Limitations). A V-JEPA2 world-model
encoder---even its native predictor---is weak for cross-instance semantic retrieval.

\begin{table*}[t]\centering\footnotesize
\caption{Full baselines and method variants (fair test, $n{=}2{,}430$; R@$k$/MRR, \%). Encoder frozen unless
noted (LoRA). ``$\rightarrow$X'' is the retrieval target space. $^\dagger$rows are $5$-seed means; \emph{all
other LFTR rows, including the two cross-space fusion rows, are seed-0 single runs from an independent
all-cutoff snapshot}, so they can differ from the $5$-seed means \emph{and} from the seed-0
encoder$\times$target grid (Table~\ref{tab:matrix}) by run-to-run noise (typically $0.4$--$1.5$pp);
\textbf{where they conflict, the $5$-seed means (Tables~\ref{tab:main},~\ref{tab:fusion}) and the grid
(Table~\ref{tab:matrix}) are canonical.} Crosswalk: this snapshot's $\rightarrow$BGE $12.8$ / $\rightarrow$DINOv2
$13.7$ / DINOv2$+$BGE $17.41$ / DINOv2$+$BGE$+$Gemma $18.11$ correspond to $5$-seed $11.3$ / $12.8{\pm}0.7$ /
$16.1{\pm}1.0$ / $17.6{\pm}0.8$ respectively. The overall \emph{best} configuration is the $4$-way
BGE$+$Gemma$+$DINOv2$+$LingBot ($18.6{\pm}0.6$, $5$-seed; Table~\ref{tab:fusion}), not shown here as we lack
its all-cutoff seed-0 breakdown.}
\label{tab:full}
\begin{tabular}{llrrrrrr}
\toprule
Group & Setup & R@1 & R@5 & R@10 & R@20 & R@50 & MRR \\
\midrule
Ceiling & Oracle (true state $\rightarrow$ BGE) & 23.95 & 52.96 & 67.82 & 80.78 & 89.63 & 37.62 \\
\midrule
MLLM & M3 (Qwen3-VL-32B) & 5.19 & 11.93 & 15.43 & 19.63 & 26.26 & 8.59 \\
MLLM & M3 (LLaVA-1.6-7B) & 1.69 & 4.20 & 6.91 & 8.85 & 11.73 & 3.11 \\
MLLM & M3 (LLaVA-NeXT-Video-7B) & 1.07 & 2.76 & 4.32 & 5.88 & 7.33 & 2.03 \\
MLLM & M3 (LLaVA-NeXT-Video-34B) & 1.48 & 4.16 & 6.54 & 8.81 & 11.98 & 2.98 \\
MLLM & M3 (27B coding VLM) & 1.32 & 4.57 & 6.50 & 8.52 & 11.24 & 2.98 \\
\midrule
Naive & Random & 0.0 & 0.0 & 0.04 & 0.04 & 0.12 & 0.01 \\
Naive & Question-only (text $\rightarrow$ BGE) & 1.60 & 5.51 & 8.48 & 11.52 & 16.79 & 3.79 \\
Naive & Text-only predictor ($7$B, states $\rightarrow$ LLM) & 2.76 & 6.50 & 8.77 & 11.48 & 15.43 & 4.69 \\
Naive & Text-only predictor ($32$B, states $\rightarrow$ LLM) & 3.91 & 9.30 & 12.51 & 16.17 & 20.66 & 6.61 \\
Naive & kNN-future (non-parametric) & 1.48 & 4.53 & 5.64 & 7.86 & 10.99 & 2.97 \\
Naive & Persistence (Qwen last-frame) & 3.91 & 10.82 & 13.95 & 18.02 & 25.06 & 7.40 \\
Naive & Persistence (DINOv2 last-frame) & 3.05 & 8.48 & 11.48 & 14.65 & 20.21 & 5.66 \\
World-model & V-JEPA2 native predictor & 1.44 & 3.83 & 4.57 & 5.19 & 7.08 & 2.56 \\
\midrule
Zero-shot & Qwen-VL $\rightarrow$ Qwen & 2.59 & 7.41 & 11.11 & 14.03 & 20.41 & 5.23 \\
Zero-shot & SigLIP $\rightarrow$ SigLIP & 2.30 & 7.74 & 9.75 & 13.00 & 19.51 & 4.85 \\
Zero-shot & InternVideo2 $\rightarrow$ InternVideo2 & 1.52 & 7.12 & 9.84 & 13.42 & 19.38 & 4.53 \\
Zero-shot & LanguageBind $\rightarrow$ LanguageBind & 1.98 & 6.83 & 10.16 & 13.70 & 19.75 & 4.64 \\
Zero-shot & DINOv2 $\rightarrow$ DINOv2 & 1.07 & 5.47 & 7.37 & 10.00 & 15.93 & 3.22 \\
Zero-shot & LingBot-Vision $\rightarrow$ LingBot & 1.40 & 4.94 & 7.41 & 10.29 & 15.06 & 3.34 \\
Zero-shot & CLIP $\rightarrow$ CLIP & 1.77 & 5.14 & 6.01 & 8.40 & 12.30 & 3.42 \\
Zero-shot & V-JEPA2 $\rightarrow$ V-JEPA2 & 1.03 & 3.70 & 4.36 & 5.68 & 7.41 & 2.20 \\
\midrule
Trained-head & Qwen-VL $\rightarrow$ BGE & 3.70 & 11.60 & 15.14 & 18.19 & 24.69 & 7.49 \\
Trained-head & Qwen-VL $\rightarrow$ DINOv2 & 4.28 & 10.95 & 15.27 & 20.21 & 25.35 & 7.72 \\
Trained-head & Qwen-VL $\rightarrow$ LanguageBind & 4.53 & 10.33 & 13.13 & 17.65 & 24.86 & 7.40 \\
Trained-head & Qwen-VL $\rightarrow$ InternVideo2 & 3.17 & 8.44 & 11.81 & 16.26 & 24.40 & 6.07 \\
Trained-head & DINOv2 $\rightarrow$ BGE & 3.87 & 9.79 & 12.63 & 15.35 & 19.88 & 6.83 \\
Trained-head & CLIP $\rightarrow$ DINOv2 & 2.80 & 7.98 & 10.00 & 13.79 & 19.63 & 5.41 \\
Trained-head & V-JEPA2 $\rightarrow$ BGE & 3.29 & 6.09 & 8.07 & 10.62 & 14.44 & 4.88 \\
LoRA & DINOv2 & 2.10 & 7.74 & 10.12 & 12.18 & 16.58 & 4.73 \\
LoRA & V-JEPA2 & 2.22 & 6.34 & 9.18 & 11.28 & 17.24 & 4.38 \\
LoRA & Qwen-VL (sub-4k) & 2.10 & 6.26 & 8.27 & 11.19 & 15.72 & 4.10 \\
\midrule
LFTR & Qwen-VL $\rightarrow$ BGE & 3.83 & 12.80 & 16.50 & 19.71 & 25.35 & 7.83 \\
LFTR & Qwen-VL $\rightarrow$ EmbeddingGemma$^\dagger$ & 4.68 & 12.32 & 16.31 & 19.87 & 25.61 & 8.36 \\
LFTR & Qwen-VL $\rightarrow$ LingBot-Vision$^\dagger$ & 5.83 & 13.56 & 17.33 & 21.50 & 28.34 & 9.62 \\
LFTR & Qwen-VL $\rightarrow$ SigLIP$^\dagger$ & 4.47 & 10.58 & 14.07 & 18.53 & 25.74 & 7.59 \\
LFTR & Qwen-VL $\rightarrow$ LanguageBind$^\dagger$ & 4.59 & 11.03 & 14.49 & 18.34 & 24.30 & 7.80 \\
LFTR & Qwen-VL $\rightarrow$ InternVideo2$^\dagger$ & 4.44 & 11.43 & 15.17 & 19.64 & 26.83 & 8.01 \\
LFTR & Qwen-VL $\rightarrow$ DINOv2 & 5.60 & 13.70 & 17.04 & 21.19 & 27.90 & 9.47 \\
LFTR & InternVideo2 $\rightarrow$ DINOv2 & 5.76 & 12.59 & 15.84 & 18.72 & 24.57 & 8.96 \\
LFTR & SigLIP $\rightarrow$ DINOv2 & 4.77 & 11.73 & 15.39 & 18.60 & 24.90 & 8.19 \\
LFTR & LingBot-Vision $\rightarrow$ DINOv2 & 4.07 & 9.75 & 12.10 & 15.23 & 21.15 & 6.90 \\
LFTR & LanguageBind $\rightarrow$ DINOv2 & 2.63 & 7.61 & 11.23 & 14.86 & 20.33 & 5.32 \\
LFTR & SigLIP $\rightarrow$ BGE & 3.66 & 9.55 & 12.76 & 15.51 & 20.08 & 6.64 \\
LFTR & both-cond (Qwen $\rightarrow$ BGE) & 3.95 & 12.30 & 15.72 & 18.89 & 23.13 & 7.87 \\
LFTR & cross-space (DINOv2$+$BGE) & 6.50 & 17.41 & 21.65 & 25.39 & 31.52 & 11.62 \\
\textbf{LFTR} & \textbf{cross-space (DINOv2$+$BGE$+$Gemma)} & \textbf{8.11} & \textbf{18.11} & \textbf{21.60} & \textbf{25.56} & \textbf{31.19} & \textbf{12.96} \\
\bottomrule
\end{tabular}
\end{table*}

\section{Encoder $\times$ Target-Space Matrix}
\label{app:matrix}
\textbf{LFTR is encoder-agnostic (the prefix encoder is a swappable, frozen choice).} Table~\ref{tab:matrix}
is the \emph{full} $8{\times}10$ grid --- \emph{eight} diverse frozen prefix encoders (a multimodal
LLM Qwen-VL, two video--language foundation models InternVideo2/LanguageBind, two image--text encoders
SigLIP/CLIP, a self-supervised image encoder DINOv2, a self-supervised video/world-model V-JEPA2, and a
boundary-SSL encoder LingBot) $\times$ \emph{ten} target spaces (the same eight visual encoders plus the two
text-state spaces BGE/Gemma), all $80$ cells trained under one config (LFTR, R@5, seed-0). Every visual
encoder thus appears as \emph{both} a prefix row and a target column. LFTR trains and works with \emph{every}
encoder; it is not tied to Qwen-VL. The prefix-encoder ranking is Qwen-VL $>$ InternVideo2 $>$ SigLIP $>$
DINOv2 $>$ LingBot $\approx$ LanguageBind $>$ CLIP $>$ V-JEPA2: the multimodal LLM is strongest (it already
fuses vision and language) --- \emph{best for $7$ of $10$ target spaces} --- a strong \emph{video} foundation
model (InternVideo2) and a strong image--text encoder (SigLIP) are close behind, and a world-model encoder
(V-JEPA2) is \emph{not} automatically a good retrieval encoder. Qwen-VL is thus our default because it is
best, not because the method requires it. On the target axis, the boundary-SSL \emph{LingBot-Vision} space is the
strongest column (best target; Table~\ref{tab:bakeoff}) --- so good that a
\emph{DINOv2} prefix retrieving into it ($13.6$) matches the best Qwen-VL cell --- followed by the semantic
(Gemma/BGE) and DINOv2 spaces, while a contrastive-visual (SigLIP, InternVideo2) or VLM (Qwen) target is
weaker, and CLIP and especially V-JEPA2 are the \emph{weakest} targets ($\le 9.3$ / $\le 6.1$ from any
prefix). The two axes are thus decoupled: Qwen-VL is the best \emph{prefix}, LingBot the best \emph{target}.

\begin{table*}[t]\centering\small
\caption{\textbf{Full encoder $\times$ target-space grid} (LFTR R@5, $8$ frozen prefix encoders
$\times$ $10$ target spaces, one consistent config, seed-0). Rows: prefix encoder; columns: retrieval target
(BGE/Gemma text-state $51$k; DINOv2/SigLIP/IntVid2/Qwen/LingBot/CLIP/V-JEPA2/LangBind visual $218$k). All $8$
visual encoders appear as \emph{both} a prefix row and a target column (the grid is not symmetric---e.g.\
DINOv2$\rightarrow$LingBot $13.6$ vs.\ LingBot$\rightarrow$DINOv2 $9.8$); all
$8{\times}10{=}80$ cells are filled. Best prefix per column in \textbf{bold}. LFTR trains with \emph{every}
frozen encoder; Qwen-VL is the best \emph{prefix} for $7/10$ targets, while LingBot is the best \emph{target}
(its column is high for \emph{every} prefix --- $10.5$--$13.7$ from any decent one --- reached even from a
DINOv2 prefix). Read down the columns: LingBot/DINOv2/Gemma are the strongest \emph{targets}; CLIP and
especially V-JEPA2 are the weakest (their columns collapse regardless of prefix), confirming the target space
is the binding constraint, not the prefix.}
\label{tab:matrix}
\resizebox{\textwidth}{!}{%
\begin{tabular}{lcccccccccc}
\toprule
Prefix enc. & $\rightarrow$BGE & $\rightarrow$Gemma & $\rightarrow$DINOv2 & $\rightarrow$SigLIP & $\rightarrow$IntVid2 & $\rightarrow$Qwen & $\rightarrow$LingBot & $\rightarrow$CLIP & $\rightarrow$V-JEPA2 & $\rightarrow$LangBind \\
\midrule
Qwen-VL & \textbf{11.3} & \textbf{13.2} & \textbf{13.3} & \textbf{11.5} & 10.6 & \textbf{12.1} & \textbf{13.7} & \textbf{9.3} & 5.8 & 10.2 \\
InternVideo2 & 9.2 & 8.4 & 12.6 & 10.2 & 10.9 & 10.1 & 12.8 & 7.6 & 5.2 & 9.6 \\
SigLIP & 9.6 & 12.1 & 11.7 & 9.8 & \textbf{11.2} & 9.7 & 11.2 & 7.4 & \textbf{6.1} & 10.7 \\
DINOv2 & 9.8 & 10.8 & 11.3 & 9.0 & 9.1 & 8.8 & 13.6 & 7.9 & 5.1 & \textbf{11.0} \\
LanguageBind & 6.8 & 9.1 & 7.6 & 8.3 & 7.1 & 8.4 & 11.7 & 4.6 & 3.8 & 7.4 \\
CLIP & 5.4 & 7.4 & 5.8 & 6.8 & 9.1 & 6.8 & 11.1 & 6.8 & 4.3 & 8.2 \\
V-JEPA2 & 6.7 & 6.2 & 5.9 & 4.3 & 5.1 & 5.0 & 7.0 & 4.2 & 3.9 & 5.4 \\
LingBot-Vision & 7.5 & 7.9 & 9.8 & 7.9 & 9.9 & 8.4 & 10.5 & 6.0 & 5.3 & 8.2 \\
\bottomrule
\end{tabular}}
\end{table*}

\textbf{Does fusing \emph{prefix} encoders help? No---a revealing asymmetry.} Cross-space fusion helps on
the \emph{target} (retrieval) side (App.~\ref{app:fusion}); we test whether it also helps on the
\emph{prefix} (encoding) side by concatenating the per-frame features of multiple encoders as the LFTR
prefix (DINOv2 target; Table~\ref{tab:prefixfuse}). It does \emph{not}: the single strongest encoder,
Qwen-VL, beats \emph{every} fused prefix, and mixing in weaker per-frame features (DINOv2, SigLIP, and
especially V-JEPA2) monotonically \emph{dilutes} it. This asymmetry is informative: the target space is a
\emph{retrieval index} that benefits from complementary views (semantic $+$ visual cover different queries),
whereas the prefix is a single \emph{input encoding} best served by the one strongest representation---adding
weaker, redundant feature dimensions only gives the frozen head more to overfit. Fusion helps where views
are complementary, not where one view already dominates.

\begin{table}[t]\centering\small
\caption{Prefix-encoder fusion (concatenated per-frame features $\rightarrow$ DINOv2 target, R@5, seed-0).
Qwen-VL alone wins; fusing prefix encoders only hurts.}
\label{tab:prefixfuse}
\begin{tabular}{lc}
\toprule
Prefix encoder(s) & R@5 \\
\midrule
Qwen-VL (single) & \textbf{13.3} \\
Qwen-VL $+$ SigLIP & 12.8 \\
Qwen-VL $+$ DINOv2 & 11.2 \\
Qwen-VL $+$ DINOv2 $+$ V-JEPA2 & 7.9 \\
Qwen-VL $+$ V-JEPA2 & 7.5 \\
DINOv2 $+$ V-JEPA2 (no Qwen-VL) & 7.2 \\
\bottomrule
\end{tabular}
\end{table}

\textbf{Best zero-shot encoder $\neq$ best target space (a target-space bake-off).} We fix the strongest
prefix encoder (Qwen-VL) and vary \emph{only} the retrieval target space, spanning a text-state space
(BGE), image encoders (SigLIP, DINOv2), and two recent \emph{video-native} video--language foundation
models (LanguageBind~\citep{zhu2024languagebind}, InternVideo2~\citep{wang2024internvideo2}), each encoding
the corpus segment as an $8$-frame clip (Table~\ref{tab:bakeoff}). Two things stand out. First, the
\emph{zero-shot retrieval ranking is essentially inverted relative to the LFTR-target ranking}: SigLIP is
the \emph{best} zero-shot retriever ($7.74$) but the \emph{worst} learned target \emph{of these five} ($10.6$), whereas DINOv2
is the \emph{worst} zero-shot retriever ($5.47$) yet the \emph{best} \emph{of the five} ($12.8$). Second, the
two newest video--language models, despite native temporal modeling and strong zero-shot ($6.8$--$7.1$), do
\emph{not} beat DINOv2 as targets ($11.0$/$11.4$). The reason is consistent: image--text/video--text
\emph{contrastive} features are optimized to match present appearance and are excellent zero-shot, but they
are a \emph{harder, less structured target} for a predictor to regress toward; self-supervised patch
features retrieve poorly zero-shot yet are the most \emph{predictable} future-state target. The clearest
case is \emph{LingBot-Vision} (``LingBot'' for short)~\citep{lingbotvision2025}, an SSL image encoder trained with masked
\emph{boundary} modeling: it is the \emph{weakest} zero-shot retriever ($4.94$, below DINOv2) yet the
\emph{best} learned target of all ($13.6{\pm}0.2$ over 5 seeds, above DINOv2's $12.8$) --- a
spatial-/boundary-aware self-supervised objective yields the most predictable target. Tellingly, the same
LingBot encoder makes only a \emph{mid-tier prefix} ($9.75$ as a prefix encoder, App.~\ref{app:matrix},
below Qwen-VL/InternVideo2/SigLIP/DINOv2): being a good \emph{target} (a predictable retrieval index) and a
good \emph{prefix} (a rich input encoding) are different properties. Choosing a target space is thus a modeling
decision about \emph{predictability}, not a race for the best off-the-shelf retriever --- and no purely
visual space, video-native or not, closes the gap to the semantic ceiling, reinforcing our
language-grounded thesis (Sec.~\ref{sec:analysis}).
\textbf{The semantic axis \emph{quality} is a lever.} Consistent with this, swapping the semantic
text-state encoder from BGE-small to the stronger \emph{EmbeddingGemma-300m}~\citep{embeddinggemma2025}
lifts the semantic space (LFTR R@5 $11.3{\to}12.3$, 5-seed) \emph{and} the headline fusion (BGE$+$Gemma$+$DINOv2
$17.6{\pm}0.8$ vs.\ BGE$+$DINOv2 $16.1{\pm}1.0$; and the best 4-way BGE$+$Gemma$+$DINOv2$+$LingBot reaches
$18.6{\pm}0.6$; App.~\ref{app:fusion}). Unlike the visual encoder---interchangeable and capped near the visual floor---the
\emph{semantic} encoder's quality directly improves retrieval, underscoring that the binding representation
in PSR is the language-aligned state.

\begin{table}[t]\centering\small
\caption{Target-space bake-off at the Qwen-VL prefix: zero-shot (prefix embedding $\rightarrow$ corpus,
no prediction) vs.\ LFTR (5-seed mean R@5). Video-native encoders encode the corpus segment as an
$8$-frame clip. $^\ddagger$For text-state targets there is no prefix-frame embedding, so zero-shot is the
question-only baseline (question text $\rightarrow$ state space, no prediction). $^\dagger$Seed-0 (from the
grid of App.~\ref{app:matrix}); all others are 5-seed means. The zero-shot order is \emph{inverted} relative
to the learned-target order, and CLIP/V-JEPA2 confirm the pattern at the low end --- the two weakest learned
targets.}
\label{tab:bakeoff}
\resizebox{\columnwidth}{!}{%
\begin{tabular}{llcc}
\toprule
Target space & Corpus modality & Zero-shot & LFTR \\
\midrule
LingBot-Vision & image (SSL, boundary) & 4.94 & \textbf{13.6} \\
DINOv2 & image (SSL) & 5.47 & 12.8 \\
EmbeddingGemma & text-state & 6.95$^\ddagger$ & 12.3 \\
InternVideo2 & video (V--L) & 7.12 & 11.4 \\
BGE & text-state & 5.51$^\ddagger$ & 11.3 \\
LanguageBind & video (V--L) & 6.83 & 11.0 \\
SigLIP & image (V--L) & \textbf{7.74} & 10.6 \\
CLIP & image (V--L) & 5.14 & 9.3$^\dagger$ \\
V-JEPA2 & video (SSL, world) & 3.70 & 5.8$^\dagger$ \\
\bottomrule
\end{tabular}}
\end{table}

\section{Ablations and Hyperparameters}
\label{app:ablation}
\textbf{Components (semantic space).} In the BGE space, removing question conditioning, hard negatives,
grounding, or the uncertainty head each moves R@5 within seed noise (Table~\ref{tab:abl}); this is the
finding that motivated the space-resolved hard-negative study (App.~\ref{app:hardneg}), where the same knob
is strongly significant in the \emph{visual} space. \textbf{Hyperparameters.} Samples $K$ and rollout depth
$H$ are within noise on a single BGE model ($K{\in}\{4,16,32,64\}$: $11.1$--$12.0$;
$H{\in}\{1,2,4,8,16\}$: $11.3$--$12.7$)---consistent with App.~\ref{app:horizon} showing the multi-step
rollout is not the performance lever. We use $K{=}32$, $H{=}8$, $\tau{=}0.04$, grounding $0.5$, transition
depth $3$, AdamW lr $3{\times}10^{-4}$, cosine schedule, $40$--$60$ epochs, batch $256$, select on val R@5.

\begin{table}[t]\centering\small
\caption{LFTR component ablation (Qwen$\rightarrow$BGE); all within $\pm0.6$ seed noise.}
\label{tab:abl}
\begin{tabular}{lcc}
\toprule
Variant & R@5 & MRR \\
\midrule
full & 11.32 & 7.48 \\
$-$ question conditioning & 11.85 & 7.72 \\
$-$ hard negatives & 11.93 & 7.94 \\
$-$ grounding & 11.44 & 7.46 \\
$-$ uncertainty head ($K{=}1$) & 11.81 & 7.86 \\
\bottomrule
\end{tabular}
\end{table}

\section{Hard Negatives: Space-Resolved}
\label{app:hardneg}
\begin{figure*}[t]\centering
\includegraphics[width=\textwidth]{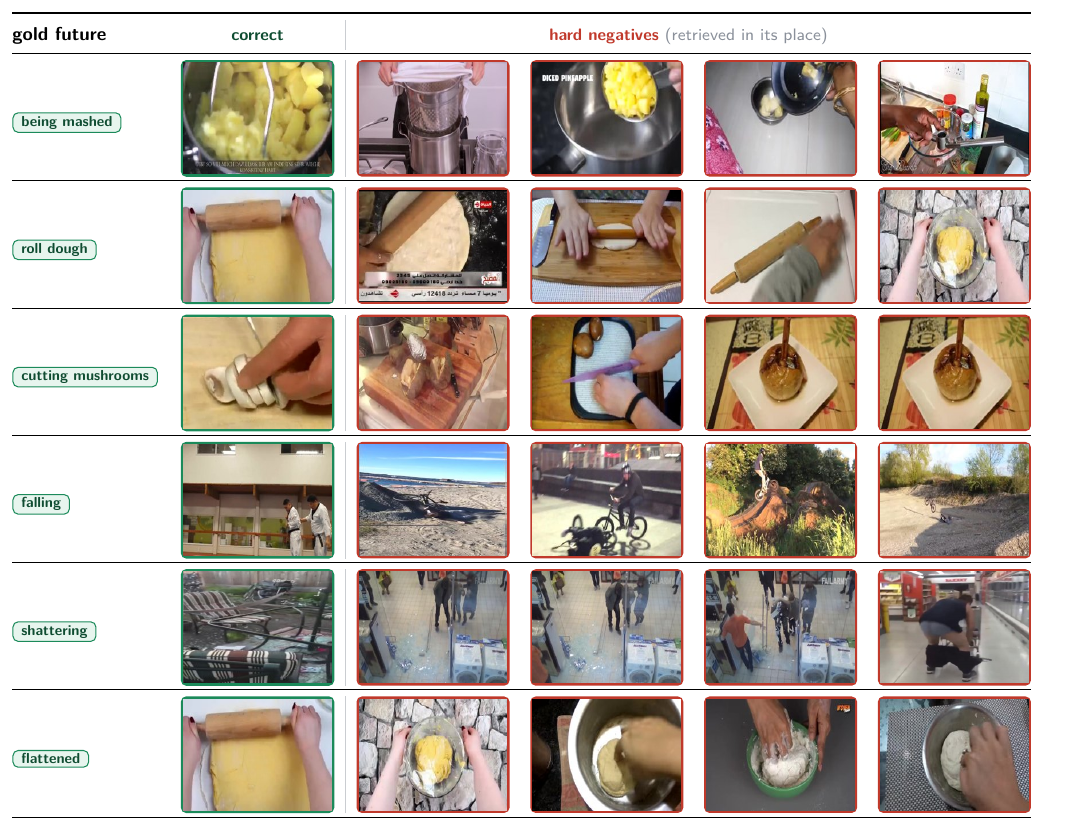}
\caption{\textbf{Positive vs.\ hard-negative future states.} Each row: the correct future-state frame
(\textcolor{green!55!black}{green}) and the \emph{hard negative} a model retrieves in its place
(\textcolor{red!62!black}{red})---a visually and semantically adjacent but \emph{wrong} future drawn from
\emph{another video}, with the BGE state-text cosine similarity that makes it confusable. Each panel is the
single middle frame (the still image encoded for training), not a full clip. These near-miss states are what
LFTR's hard-negative objective is trained to separate; mining them in the visual space (not the text space)
yields the significant gain in Table~\ref{tab:hardneg}.}
\label{fig:hardneg}
\end{figure*}

\begin{figure*}[t]\centering
\includegraphics[width=\textwidth]{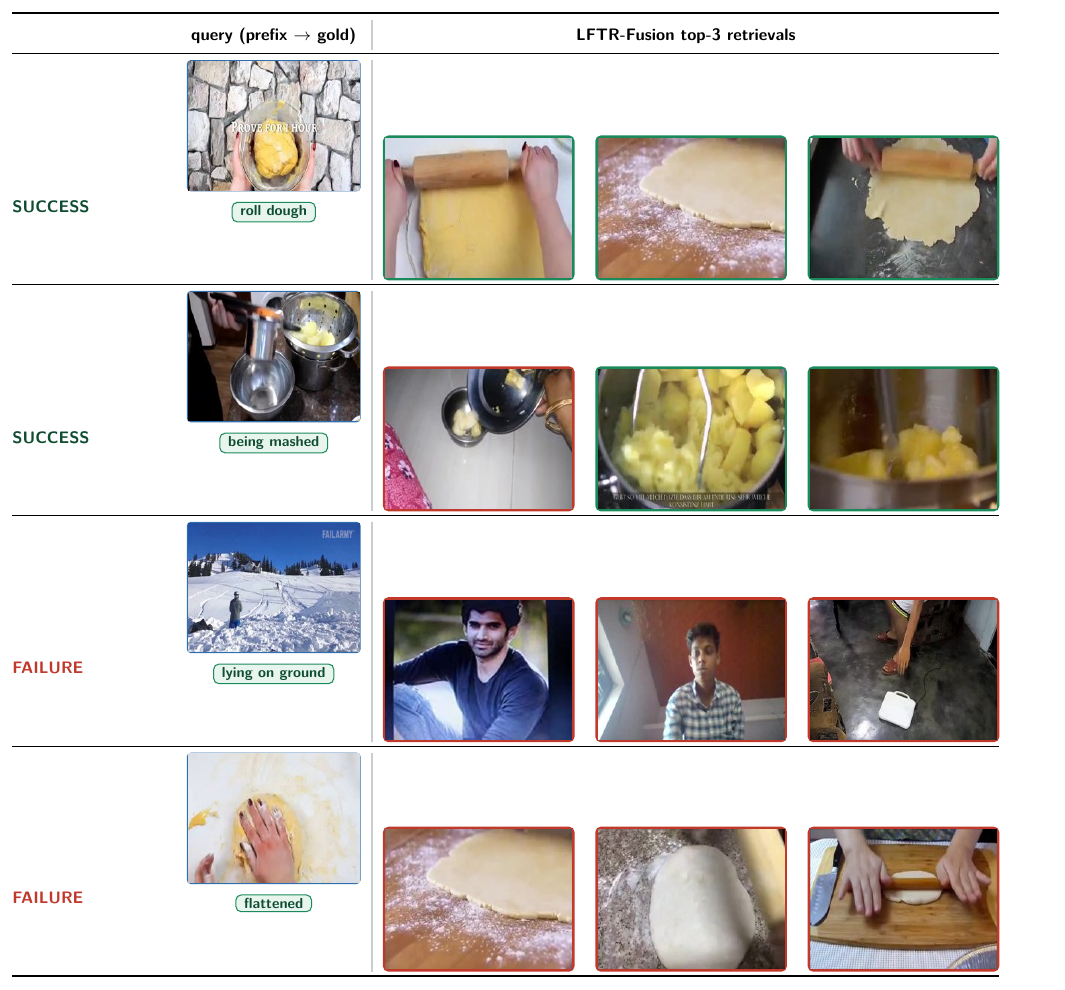}
\caption{\textbf{Positive (success, \textcolor{green!55!black}{green}) vs.\ negative (failure,
\textcolor{red!62!black}{red}).} For each query (observed prefix and gold future), LFTR-fusion's top-1
retrieval: a \emph{success} retrieves the correct future state; a \emph{failure} retrieves a \emph{hard
negative}---a semantically adjacent but wrong future state drawn from another video, rather than an
unrelated item.}
\label{fig:posneg}
\end{figure*}

Hard negatives mine embedding-similar (visually/semantically confusable) but wrong future states. In the
DINOv2 \emph{visual} space they yield a large, significant, monotone gain saturating near $128$
(Table~\ref{tab:hardneg}: $+3.9$pp R@5, $p<10^{-3}$, five seeds). In the BGE \emph{semantic}
space the same knob is neutral (Table~\ref{tab:abl}), because text states are already lexically separated.
Visual, not semantic, confusability is the binding constraint---matching the failure analysis
(Fig.~\ref{fig:posneg}), where top-1 errors are semantically adjacent states
(\emph{lying on ground}$\rightarrow$``being pushed'', BGE sim $0.61$; \emph{being grated}$\rightarrow$
``grating parmesan'', $0.69$). Fig.~\ref{fig:hardneg} makes this concrete: the correct future frame sits
among hard negatives that are near-identical scenes in a \emph{different} future state (chopping meat vs.\
cutting chicken / mincing garlic; falling into water vs.\ already submerged), which the multi-query
contrastive objective must rank apart.

\begin{table}[t]\centering\small
\caption{Hard negatives in the DINOv2 (visual) space: R@5 vs.\ number of mined embedding-similar
negatives (5 seeds). $n_{\mathrm{hard}}{=}0$ is random-negatives-only.}
\label{tab:hardneg}
\begin{tabular}{lcc}
\toprule
$n_{\mathrm{hard}}$ & R@5 & $\Delta$ vs.\ 0 (sig.) \\
\midrule
0 (random only) & $8.98{\pm}0.91$ & --- \\
32 & $11.34{\pm}0.44$ & $+2.36$ (\,$p{=}0.004$\,) \\
128 & $12.80{\pm}0.72$ & $+3.82$ (\,$p{=}2{\times}10^{-4}$\,) \\
256 & $12.92{\pm}0.50$ & $+3.94$ (\,$p{=}2{\times}10^{-4}$\,) \\
\bottomrule
\end{tabular}
\end{table}


\section{Cross-Space Fusion}
\label{app:fusion}

\begin{figure*}[t]\centering
\includegraphics[width=\textwidth]{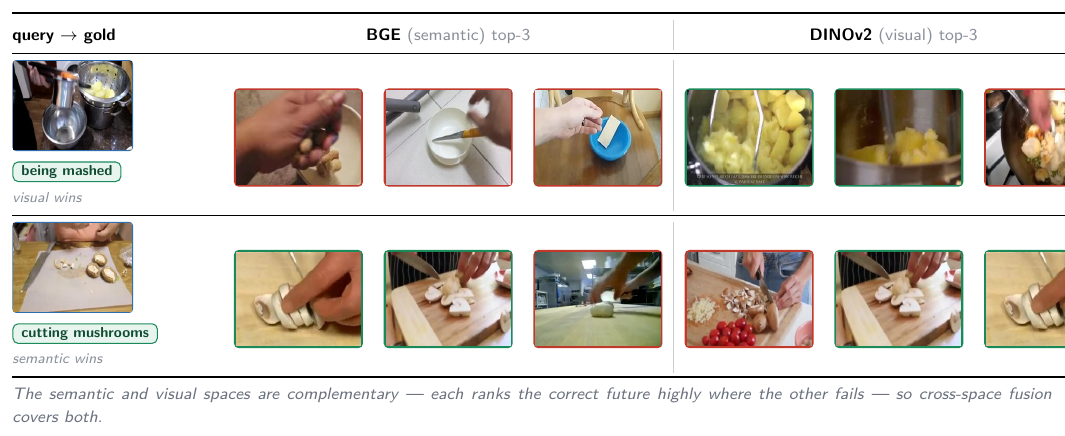}
\caption{\textbf{Why fusion works: complementarity.} One query per row (observed prefix and gold future on
the left; each method's top-1 retrieval, \textcolor{green!55!black}{green} if correct,
\textcolor{red!62!black}{red} otherwise). The semantic (BGE) and visual (DINOv2) spaces make \emph{different}
errors --- top row the visual space is right and the semantic wrong, bottom row the reverse --- so z-score
cross-space fusion recovers the correct future where either space alone fails.}
\label{fig:compl}
\end{figure*}

\begin{figure*}[t]\centering
\includegraphics[width=\textwidth]{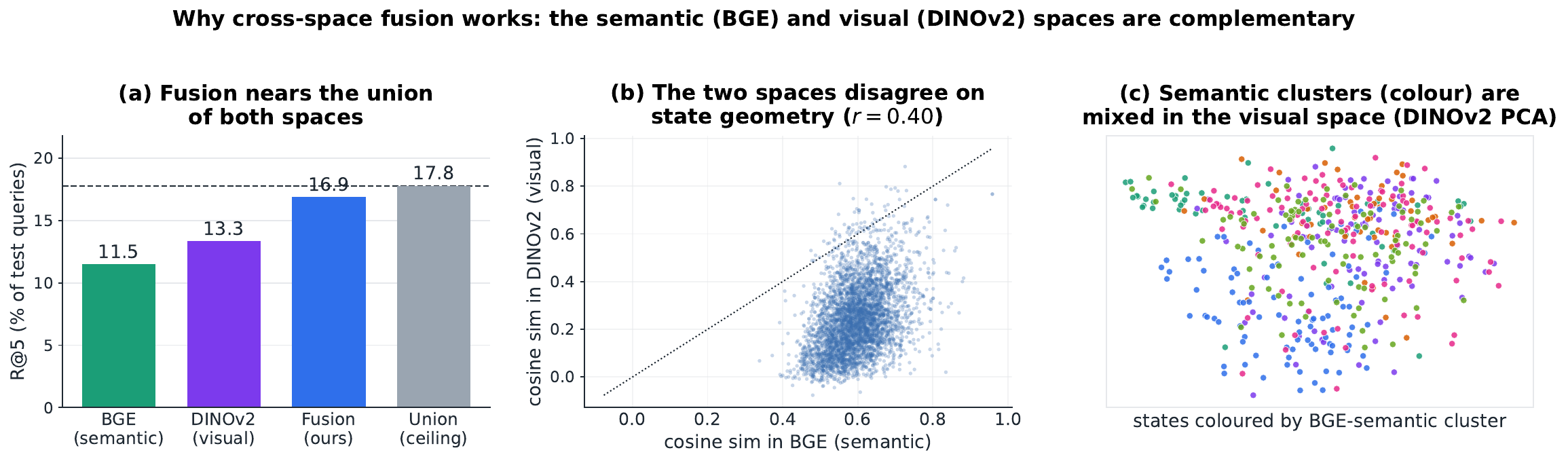}
\caption{\textbf{Why cross-space fusion works.} (a)~Complementary coverage: of the test queries, $4.4\%$
are solved only by the semantic (BGE) space and $6.3\%$ only by the visual (DINOv2) space; $z$-score fusion
($16.9$) nearly recovers their \emph{union} ($17.8$), far above either alone. (b)~The two spaces disagree on
state geometry: cosine similarity of state pairs in BGE vs.\ DINOv2 is weakly correlated ($r{=}0.40$).
(c)~States clustered by \emph{semantics} (BGE, colour) are spatially \emph{mixed} when laid out by
\emph{visual} similarity (DINOv2 PCA)---the two spaces organise the same states differently, which is the
source of the complementarity fusion exploits. Panel~(a) is a single representative run, so its absolute R@5
runs ${\sim}1$pp above the seed-averaged headline (fusion $16.1$, Table~\ref{tab:main}).}
\label{fig:fusionwhy}
\end{figure*}

\textbf{Which spaces? A full combination sweep.} We put six trained target spaces (BGE, DINOv2, SigLIP,
LanguageBind, InternVideo2, Qwen) into \emph{one} common fusion framework (each is an LFTR head on the
Qwen-VL prefix, so they are directly comparable) and $z$-score--fuse every pair, triple, and the full
six-way combination (Table~\ref{tab:fusion}); we additionally evaluate the stronger EmbeddingGemma semantic
space and the boundary-SSL LingBot visual space, and report all headline fusions over $5$ seeds ($^{**}$
denotes a significant per-seed paired improvement over BGE$+$DINOv2). Three patterns emerge. (i) \textbf{The semantic$\times$visual axis is what matters}: \emph{every}
top combination pairs the semantic BGE space with a visual one, and \emph{every} combination \emph{without}
BGE plateaus $\sim$$3$pp lower (e.g., DINOv2$+$SigLIP$+$IntVid2 $14.7$). (ii) \textbf{The specific visual
encoder is interchangeable}: BGE$+$DINOv2/SigLIP/LangBind/InternVideo2 all land $16.9$--$17.4$, with DINOv2
best. (iii) \textbf{More spaces help only if each adds a complementary \emph{strong} axis, not a redundant
one.} Piling on a \emph{redundant same-type} encoder is useless---a second contrastive-visual space on top of
DINOv2 (BGE$+$DINOv2$+$SigLIP $17.4$) or the full six-way fusion ($17.0$) does not beat the best BGE$+$visual
triple. \emph{But} the best encoder of each \emph{distinct} family \emph{does} stack: the strongest semantic
space (EmbeddingGemma) and two \emph{differently-supervised} visual encoders (DINOv2 patch-SSL $+$ LingBot
boundary-SSL) each contribute, and combining all four---BGE$+$Gemma$+$DINOv2$+$LingBot---is the best
configuration at R@5 $\mathbf{18.6{\pm}0.6}$ over five seeds, a significant $+2.5$pp over the $2$-space
BGE$+$DINOv2 (paired $t$, $p{=}0.002$; every intermediate combo marked $^{**}$ in Table~\ref{tab:fusion} is
also significant). So the rule is \emph{complementarity, not count}: adding a space helps exactly when it
organizes states differently (a new semantic encoder, or a visual encoder with a different SSL objective),
and not when it merely duplicates an axis already covered. Fig.~\ref{fig:compl} shows this per query --- one
space is right exactly where the other is wrong --- and Fig.~\ref{fig:fusionwhy} makes the mechanism concrete: BGE and
DINOv2 solve \emph{different} queries ($4.4\%$ BGE-only $+$ $6.3\%$ DINOv2-only; this coverage analysis uses a
single run at fusion $16.9$ --- a different seed from the $16.75$ run used for the paired test above, both
within the $16.1{\pm}1.0$ band --- and it nearly matches their union $17.8$), their per-pair state-similarity
is weakly correlated ($r{=}0.40$), and
semantic clusters are spatially \emph{mixed} in the visual space---the two views organize states
differently, so combining them adds information.
\textbf{Significance.} Five seed-paired fusions give R@5 $16.06{\pm}0.95$ ($=16.1$ in Table~\ref{tab:main})
vs.\ M3 $11.93$ ($+4.13$pp). For a clean, selection-free test we also compare fusion and M3 \emph{per query} on
the identical set of all $n{=}2{,}430$ test queries, using \emph{one representative fusion run} (R@5 $16.75$,
within the $16.1{\pm}1.0$ seed band) against M3's deterministic (temperature-$0$) prediction. A paired
bootstrap ($10^4$ resamples over queries; this tests query-sampling variance, not seed variance --- M3 is a
single deterministic decode) gives $\Delta$R@5 $=+4.81$pp, 95\% CI $[+3.00,+6.58]$, two-sided $p<10^{-4}$;
McNemar's exact test on the discordant pairs (fusion right / M3 wrong $=305$; M3 right / fusion wrong $=188$)
gives $p{=}1.5{\times}10^{-7}$. Both agree fusion is a strongly significant improvement.
\textbf{The win is not confined to the easy near horizon.} Stratifying the same paired bootstrap by horizon,
fusion beats M3 significantly at $h1$ ($+3.8$pp, $p{=}0.005$), $h2$ ($+9.9$pp, $p{=}10^{-4}$ --- the
\emph{largest} margin, exactly where the mid-horizon future first becomes multi-modal), and $h3$ ($+4.5$pp,
$p{=}0.04$); at $h4$/$h5$ the tiny buckets (fusion$\cap$M3 common set, $n{=}109/77$) leave it non-significant
($-0.9$/$+6.5$pp). So the
advantage holds precisely where forecasting starts to bite, not only at $h1$. Fusion also beats the
strong \emph{persistence} baseline by a \emph{larger} margin ($16.1$ vs.\ $10.8$, $+5.3$pp) than it beats M3,
so ``a method must predict'' holds under the same test. Single-space LFTR over M3 is within seed noise
($+1.65$pp, $p{=}0.063$), so \emph{fusion} is the robust win. These are the paper's two pre-registered
comparisons (fusion vs.\ M3, fusion vs.\ persistence); the many exploratory per-combination tests in
Tables~\ref{tab:hardneg},~\ref{tab:fusion} are reported with raw $p$-values and should be read as exploratory
(no multiple-comparison correction applied).

\begin{table}[t]\centering\small
\caption{Fusion combination sweep (R@5, $n{=}2{,}430$). Fusion rows with $\pm$ are $5$-seed
mean$\pm$std with a per-seed paired $t$-test vs.\ BGE$+$DINOv2; single-space rows and the redundant-sweep
block are seed-0. \textbf{Complementary} strong axes stack (best semantic Gemma $+$ two differently-supervised
visual encoders DINOv2/LingBot), giving the best config; \textbf{redundant} same-type encoders (a 2nd
contrastive-visual space, or the whole six-way) do \emph{not}.}
\label{tab:fusion}
\resizebox{\columnwidth}{!}{%
\begin{tabular}{lc}
\toprule
Combination & R@5 \\
\midrule
\multicolumn{2}{l}{\emph{Single spaces (seed-0)}} \\
LingBot / DINOv2 / EmbeddingGemma & 13.6 / 13.3 / 13.3 \\
BGE & 11.5 \\
SigLIP / IntVid2 / LangBind / Qwen & 11.1 / 10.8 / 10.3 / 10.2 \\
\midrule
\multicolumn{2}{l}{\emph{Fusions (5-seed)}} \\
BGE $+$ DINOv2 & $16.1{\pm}1.0$ \\
BGE $+$ LingBot & $16.7{\pm}0.6$ \\
Gemma $+$ LingBot & $17.2{\pm}0.7$ \\
BGE $+$ DINOv2 $+$ LingBot & $17.5{\pm}0.8^{**}$ \\
BGE $+$ Gemma $+$ DINOv2 & $17.6{\pm}0.8^{**}$ \\
BGE $+$ Gemma $+$ LingBot & $17.8{\pm}0.5^{**}$ \\
\textbf{BGE $+$ Gemma $+$ DINOv2 $+$ LingBot} & $\mathbf{18.6{\pm}0.6}^{**}$ \\
\midrule
\multicolumn{2}{l}{\emph{Redundant same-type (seed-0, no gain)}} \\
BGE $+$ DINOv2 $+$ SigLIP & 17.4 \\
All six spaces & 17.0 \\
DINOv2 $+$ SigLIP $+$ IntVid2 (no semantic) & 14.7 \\
\bottomrule
\end{tabular}}
\end{table}

\section{Appearance Floor Across Visual Spaces}
\label{app:oraclevis}
The central ceiling claim---that the future is \emph{not} matchable by raw appearance across instances, only
by semantic state---rests on the independent observed-future-frame oracle (query $=$ the object's own future
frame, retrieve cross-instance GT). A skeptic could object that its low score ($5.7$) reflects a poor choice
of \emph{one} visual space (DINOv2) or \emph{one} frame. Table~\ref{tab:oraclevis} repeats the oracle in
every visual space and with a multi-frame query. It stays near the floor everywhere---$3.0$ (CLIP) to $6.6$
(SigLIP)---and averaging all of the query's own future frames does not help. Even the best visual oracle
($6.6$) is roughly $8\times$ below the semantic ceiling ($53$), so ``cannot solve by appearance'' is a
property of the \emph{task}, not of a particular encoder or single-frame query.

\begin{table}[t]\centering\small
\caption{Independent observed-future-frame oracle (query $=$ the object's \emph{own} future frame $\rightarrow$
retrieve cross-instance GT; $95\%$ bootstrap CIs), across every visual space and with a multi-frame (mean of
all self future frames) query. Appearance is near the floor everywhere, far below the semantic ceiling ($53$).}
\label{tab:oraclevis}
\resizebox{\columnwidth}{!}{%
\begin{tabular}{lcc}
\toprule
Query space ($=$ own future frame) & R@5 & 95\% CI \\
\midrule
DINOv2 & 5.7 & $[4.4,7.0]$ \\
SigLIP & 6.6 & $[5.3,8.1]$ \\
CLIP & 3.0 & $[2.1,4.0]$ \\
LingBot-Vision & 6.5 & $[5.1,7.9]$ \\
\midrule
DINOv2 (multi-frame) & 5.7 & $[4.4,7.0]$ \\
LingBot-Vision (multi-frame) & 6.5 & $[5.2,7.9]$ \\
\midrule
\emph{Semantic ceiling (BGE text oracle)} & \emph{53.0} & --- \\
\bottomrule
\end{tabular}}
\end{table}

\section{Horizon, Ceiling, and the Rollout}
\label{app:horizon}
\begin{figure}[t]\centering
\includegraphics[width=\columnwidth]{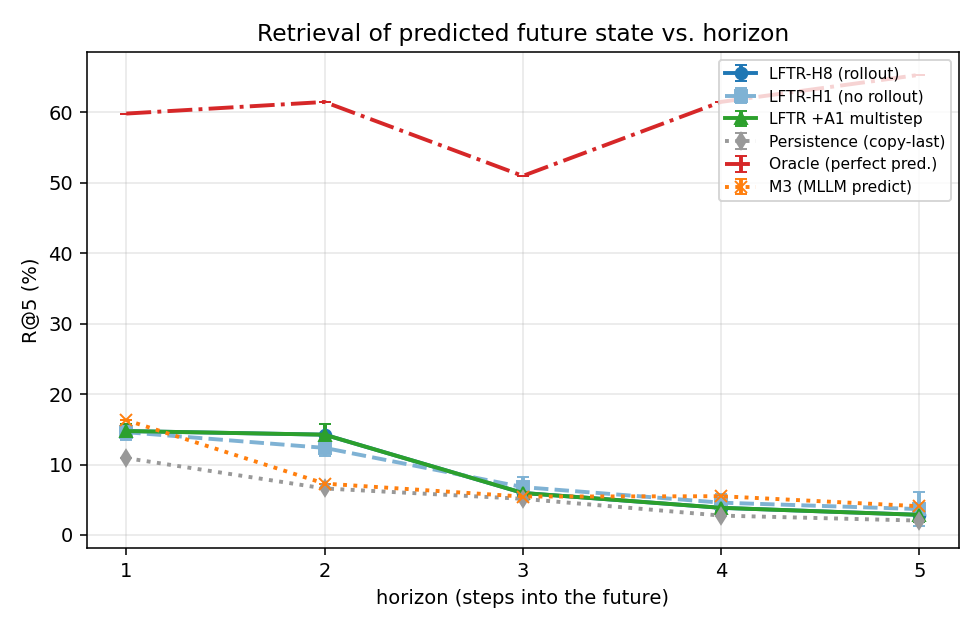}
\caption{R@5 vs.\ horizon. The Oracle (perfect prediction, red) is flat $\sim$$55\%$; every learned method
--- LFTR with the full rollout ($H{=}8$), without it ($H{=}1$), with intermediate supervision, plus
persistence and the MLLM (M3) --- collapses beyond $h{=}2$ and the rollout variants overlap. The gap to the
flat Oracle is prediction error, and the rollout mechanism does not close it.}
\label{fig:horizon}
\end{figure}

The Oracle (perfect future-state) is flat across horizon in both spaces (main Table~\ref{tab:ceiling}),
yet every learned predictor collapses past $h{=}2$ (Fig.~\ref{fig:horizon}). We isolate the rollout:
the full $H{=}8$ trajectory is statistically indistinguishable from an $H{=}1$ horizon-conditioned read
(overall $12.80{\pm}0.72$ vs.\ $12.40{\pm}1.02$, $p{=}0.54$; n.s.\ at every horizon), and intermediate-
horizon supervision does not help at any weight (Table~\ref{tab:a1}). Per-horizon test support is
$n{=}1356/576/312/109/49$ for $h{=}1..5$. Thus far-horizon \emph{prediction}, not retrieval or the rollout
mechanism, is the bottleneck.

\begin{table}[t]\centering\small
\caption{Intermediate-horizon supervision (A1) dose, long-horizon ($h{\ge}3$) and overall R@5 (5 seeds).}
\label{tab:a1}
\begin{tabular}{lcc}
\toprule
Weight & $h{\ge}3$ R@5 & overall R@5 \\
\midrule
0 (H8 baseline) & $5.15{\pm}0.31$ & $12.80{\pm}0.72$ \\
0.5 & $5.15{\pm}0.31$ & $12.80{\pm}0.73$ \\
1.0 & $5.40{\pm}0.48$ & $12.77{\pm}0.72$ \\
3.0 & $5.57{\pm}0.49$ ($p{=}0.19$) & $13.01{\pm}0.51$ \\
\bottomrule
\end{tabular}
\end{table}

\section{Rollout: A Rich but Under-Exploited Trajectory}
\label{app:rolldistinct}
The rollout is not an accuracy lever (App.~\ref{app:horizon}), but it is a \emph{structured} predictor with
a distinctive, if untapped, property. \textbf{(i) Single-pass, any-horizon.} One rollout emits the full
future trajectory $(\mu_1,\dots,\mu_H)$, so a shared single early prefix---queried at many horizons and
question types---is answered by \emph{reading} different steps of \emph{one} forward pass, rather than
recomputing per query. \textbf{(ii) The trajectory is rich.} Table~\ref{tab:rolldistinct} varies only
\emph{which} step of a fixed trained model is read. A horizon-agnostic \emph{fixed step 1} read underperforms
the default \emph{matched step $h$} read, confirming the trajectory is horizon-ordered. Crucially, an
\emph{oracle} that reads the best step \emph{per query} reaches R@5 $17.7$ (DINOv2) / $16.0$ (BGE) vs.\
$13.4$ / $11.2$ for the matched read---a large $+4.3$/$+4.8$pp head-room that \emph{replicates across both
retrieval spaces}. \textbf{(iii) The signal is per-query and hard to harvest.} A validation-learned
per-horizon step calibration adds $\approx0$pp (the best step is query-, not horizon-, specific), and a
multi-hypothesis read (max-similarity over \emph{all} steps at once) helps only at far horizons in the
visual space ($h{\ge}3$: $7.2$ vs.\ matched $5.1$, $+2.1$pp; near horizons and BGE: neutral-to-worse). So
the rollout demonstrably encodes retrievable future states across its steps, but converting that latent
richness into a robust, oracle-free read rule is an open problem the benchmark surfaces---a concrete future
direction rather than a solved contribution.

\begin{table}[t]\centering\small
\caption{Reading a fixed trained $H{=}8$ model different ways (R@5). Only the read strategy changes. The
\emph{oracle best step} (per query) reveals $+4.3$/$+4.8$pp of untapped trajectory signal in both spaces;
simple oracle-free rules (calibration, multi-hypothesis union) recover little of it on aggregate.}
\label{tab:rolldistinct}
\begin{tabular}{lcc}
\toprule
Read strategy & DINOv2 & BGE \\
\midrule
Fixed step 1 (horizon-agnostic) & 12.2 & 10.6 \\
Matched step $h$ (default) & 13.4 & 11.2 \\
Val-calibrated step & 13.6 & 11.2 \\
Multi-hypothesis (union of steps) & 12.7 & 7.6 \\
\midrule
\emph{Oracle} best step (upper bound) & \textbf{17.7} & \textbf{16.0} \\
\bottomrule
\end{tabular}
\end{table}

\section{Difficulty Tiers and Generalization}
\label{app:tiers}
\begin{figure}[t]\centering
\includegraphics[width=\columnwidth]{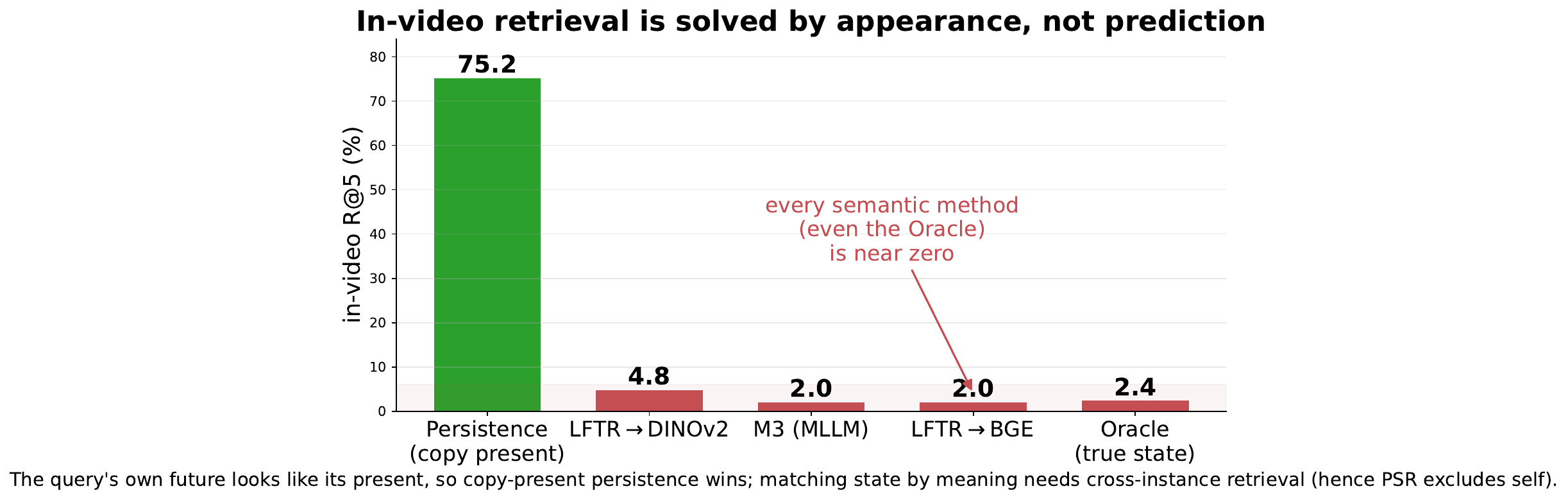}
\caption{On the in-video (self) tier, copy-the-present persistence reaches R@5 $75.2$ while every
\emph{semantic} method --- even the Oracle given the true state --- is near zero. In-video is solved by
appearance, so PSR excludes self and forces cross-instance retrieval.}
\label{fig:invideo}
\end{figure}

\textbf{In-video (self).} Appearance \emph{persistence} dominates (R@5 $75.2$) while every semantic method,
\emph{including the Oracle}, scores $\le 5$ (Fig.~\ref{fig:invideo})---in-video retrieval is solved by
visual instance-matching, not prediction, which is why the main pools exclude self.
\textbf{In- vs.\ cross-dataset.} Cross-dataset R@5 is $\approx 8.4$ for both LFTR variants (the honest
generalization gap), well below in-dataset ($13$--$17$); DINOv2's advantage is mostly in-dataset
($17.2$ vs.\ $8.4$).

\section{Robustness to GT Definition}
\label{app:gtrobust}
Across type2-all / type2-consensus / type1-all the method ranking is stable (LFTR$\rightarrow$DINOv2
$13.7/12.3/13.4$; M3 $11.9/11.8/9.9$; Oracle $53.0/51.0/38.1$)---results are not an artifact of one GT
definition. Type1 (subject$+$predicate) is stricter, lowering the ceiling to $38$.

\section{Corpus-Size Ablation}
\label{app:corpus}
As the corpus grows $20$k$\rightarrow$$218$k all methods degrade monotonically, but LFTR$\rightarrow$DINOv2
stays on top and \emph{widens} its margin over M3 ($20$k: $18.4$ vs.\ $18.2$; $218$k: $13.5$ vs.\ $11.9$;
all values here are single seed-0 runs, so the $218$k LFTR$\rightarrow$DINOv2 $13.5$ is within run noise of the
canonical $12.8{\pm}0.7$/$13.7$-seed-0), i.e.\ it degrades more gracefully. The semantic-text oracle falls
$78.1\rightarrow52.9$; the DINOv2-last-frame persistence variant $14.9\rightarrow8.5$ (the Qwen-last-frame
headline persistence is $10.8$ at $218$k).

\section{Image Modality}
\label{app:image}
On an image-only corpus (VG $+$ MIT-States, $333$k items) retrieval is hard even for the semantic-text
oracle (R@5 $8.9$), versus $49.7$ for the \emph{same} oracle restricted to this appendix's image-eligible
query subset on the video corpus (lower than the main-benchmark $53.0$ because this is a harder, smaller
subset). The drop to $8.9$ (Table~\ref{tab:image}) has two causes: a vocabulary/domain mismatch between
video-derived query states and image-corpus labels, and---more fundamentally---that many PSR states are
\emph{dynamic} (\emph{falling}, \emph{being pushed}), which a \emph{static} image captures poorly (VG and
MIT-States depict static scenes) whereas a \emph{video} segment shows the motion. Images therefore match
near-static \emph{end}-states (\emph{shattered}, \emph{sliced}) far better than mid-action ones---an open
cross-modal challenge of the benchmark.

\begin{table}[t]\centering\small
\caption{\textbf{Video vs.\ image corpus.} Same semantic-text oracle and the \emph{same} image-eligible query
subset, retrieving from each corpus. The oracle drops from $49.7$ (video) to $8.9$ (image): retrieving image
evidence for a video-derived future state is a hard cross-modal generalization test (vocabulary/domain
mismatch), while the modality is fully supported by the released \texttt{gt\_image} pools.}
\label{tab:image}
\begin{tabular}{lc}
\toprule
Retrieval corpus (same queries, same oracle) & R@5 \\
\midrule
Video corpus ($218$k segments) & 49.7 \\
Image corpus (VG $+$ MIT-States, $333$k) & \phantom{0}8.9 \\
\bottomrule
\end{tabular}
\end{table}

\section{Multi-Object (Q5)}
\label{app:q5}
Applied per object, LFTR$\rightarrow$DINOv2 matches its single-object end-state score (per-object R@5
$8.84$ vs.\ Q2 single $9.11$) and leads M3 ($7.67$), so the method transfers. But \emph{joint}
all-objects-correct is combinatorial ($\approx p^{K}$) and near zero even for the Oracle ($18.8\%$ joint
for two objects, collapsing with more)---the low absolute scores stem from end-state difficulty plus the
combinatorial joint metric, not method unsuitability. Joint multi-object prediction is an open regime.

\section{Efficiency}
\label{app:eff}
The LFTR head is $11.98$M parameters and inference is $0.06$\,ms/query (predict $+$ retrieve) with the encoder
frozen and features precomputed --- roughly four orders of magnitude cheaper than generating with a $32$B MLLM
per query. Table~\ref{tab:compute} gives the full budget on a single \textsc{nvidia} H200: because encoders are
frozen and their features precomputed once, training the head is cheap ($<1$ GPU-hour per run) and the whole
multi-seed, multi-space study is only a few GPU-hours; the dominant one-time cost is precomputing the corpus
feature indices, which we release so the benchmark reproduces without re-encoding.

\begin{table}[t]\centering\small
\caption{\textbf{Compute budget} (single \textsc{nvidia} H200). Encoders are \emph{frozen} and their features
precomputed once, so training the $\sim$12M head is cheap; precomputing the corpus feature indices is the
dominant one-time cost. Inference is per query with features cached.}
\label{tab:compute}
\resizebox{\columnwidth}{!}{%
\begin{tabular}{llr}
\toprule
Stage & Trained params & Compute \\
\midrule
Corpus feature precompute (one-time) & 0 (frozen)   & $2.5$\,GPU-h \\
LFTR head training (per run, $40$ ep.) & $11.98$M     & $<1$\,GPU-h \\
Full multi-seed $\times$ space sweep   & $11.98$M     & $\sim$$6$\,GPU-h \\
LoRA encoder fine-tune (per epoch)     & $\sim$$18.5$M & $\sim$$40$\,min \\
\midrule
Inference: predict $+$ retrieve        & ---          & $0.06$\,ms/query \\
\bottomrule
\end{tabular}}
\end{table}

\section{Does Unfreezing the Encoder Help?}
\label{app:unfreeze}
We probe whether LFTR is limited by its frozen prefix, its objective, or its target. \textbf{Objective/target
changes are null.} Redirecting the grounding loss from the query's GT-instance centroid to the state
prototype the Oracle uses (a natural ``align with the ceiling'' idea) changes R@5 by $-0.02$ over five seeds
($p{=}0.96$); the multi-step rollout is statistically indistinguishable from a single horizon-conditioned
read; and intermediate-horizon supervision does not help at any weight. \textbf{Only unfreezing the encoder
helps, and only slightly.} We LoRA-fine-tune the prefix encoder end-to-end through the LFTR head, with a
frozen-encoder run of the \emph{same} script as the control (Table~\ref{tab:unfreeze}). Unfreezing gives
$+0.53$ (DINOv2) and $+0.37$ (CLIP) R@5 --- consistent across two encoders but within single-seed noise
($\pm0.7$) and at $\sim$$40$\,min/epoch versus seconds/epoch for a frozen head on cached features. The frozen
prefix is therefore a \emph{mild} constraint; even a fully adaptable encoder stays $\sim$$45$pp below the
horizon-flat Oracle, so far-horizon prediction --- not the encoder, head, or loss --- is the binding
bottleneck. Frozen-LFTR is the efficient Pareto point.

\begin{table}[t]\centering\small
\caption{Frozen vs.\ LoRA end-to-end prefix encoder (same head, schedule, eval; R@5). Unfreezing helps
marginally, within seed noise, at $\sim$$10^3\times$ the training cost.}
\label{tab:unfreeze}
\begin{tabular}{lccc}
\toprule
Encoder & frozen & LoRA end-to-end & $\Delta$ \\
\midrule
DINOv2 & 10.66 & 11.19 & $+0.53$ \\
CLIP & 7.41 & 7.78 & $+0.37$ \\
\bottomrule
\end{tabular}
\end{table}

\section{Prompt Templates}
\label{app:prompts}
The exact prompts, including the multi-pass \emph{verification} prompts used for GT quality control (omitted
here for space but equally load-bearing for the reliability claims), are released with the code; we reproduce
the generation prompts verbatim below. Braces are
substituted per query.

\textbf{Text-only predictor (oracle-perception control, \S\ref{sec:analysis}).}
\begin{quote}\small\ttfamily
An object ('\{subject\}') was observed with these states over time: \{t$_1$s: state$_1$; t$_2$s: state$_2$;
\dots\}. \{question\} Answer with ONLY the object's predicted future state as a short phrase (2--5 words), no
explanation.
\end{quote}
The state list is truncated to states with timestamp $<$ the query's target time, so the answer is never
revealed.

\textbf{M3 (multimodal-LLM baseline).} Identical to the text-only prompt but the $8$ prefix frames are
prepended as images and the state transcript is omitted, so the model must read the states off the frames:
\begin{quote}\small\ttfamily
[8 prefix frames] \{question\} Answer with ONLY the object's predicted future state as a short phrase
(2--5 words), no explanation.
\end{quote}
Decoding is temperature $0$ (deterministic), \texttt{max\_tokens}$=24$; the phrase is BGE-encoded and
retrieved.

\textbf{Benchmark construction (state/question generation).} A representative template; the full loop
(object inventory, timed state trajectory, per-type question generation) is in the released
\texttt{build\_dataset} scripts:
\begin{quote}\small\ttfamily
Watch this video segment. List each visible object and its state trajectory as (state, time) pairs. Then, for
the object '\{subject\}', write a \{Q-type\} question whose answer is its \{next / end / after-$k$s /
intermediate\} state, and give that ground-truth state as a short predicate phrase.
\end{quote}

\section{Retrieval-Channel Decomposition}
\label{app:channel}
The raw method--ceiling gap ($16\to53$) mixes two distinct sub-problems, and separating them clarifies what a
better \emph{predictor} can actually close. All learned methods retrieve a \emph{free-form predicted} state
string, whereas the semantic oracle ($53$) retrieves the dataset's \emph{canonical} string; so part of the
gap is a \emph{retrieval-canonicalization} channel (how well a correct-but-differently-phrased state matches
the canonical index), not forecasting.

\textbf{Forecasting vs.\ canonicalization (clean control).} We paraphrase each query's \emph{true} state with
an LLM (reword, same meaning) and retrieve the paraphrase. This holds the \emph{content} correct and varies
only the \emph{phrasing}, so any drop from $53$ is pure canonicalization loss. The paraphrase oracle reaches
R@5 $\mathbf{29.7}$ (Table~\ref{tab:channel}). Two consequences: (i) the ceiling \emph{achievable by any
free-form forecaster} is $\sim$$30$, not $53$; (ii) the genuine \emph{forecasting} gap a better predictor can
close is therefore $\sim$$14$pp ($16\to30$), while the $30\to53$ residual is an artifact of exact-string
canonicalization --- a lever for future \emph{benchmark} design (e.g., paraphrase-robust GT indices), not for
the predictor. This sharpens rather than weakens the ``prediction is the open problem'' message: it gives a
concrete, achievable target ($30$) and isolates the learnable portion of the gap.

\textbf{Cross-model description (rough, confounded).} As a rough probe of whether the canonical vocabulary is
generator-specific, we let a \emph{non-Gemini} VLM (LLaVA-1.6-7B) describe each query's true GT segment from
scratch and retrieve; it reaches only R@5 $3.2$. We do \emph{not} lean on this number: it conflates genuine
vocabulary mismatch with a much weaker single-\emph{frame} captioner describing a state that often needs
temporal context (e.g.\ ``falling into water'' looks like ``in water'' in one frame). A clean test of
generator-specificity requires full cross-generator \emph{re-authoring} of states/questions/GT with a
different VLM family, which we leave to future work (Limitations).

\begin{table}[t]\centering\small
\caption{Retrieval-channel decomposition (R@5). Paraphrasing the \emph{true} state (content fixed, phrasing
varied) isolates the canonicalization channel; the achievable free-form ceiling is $\sim$$30$, so the genuine
forecasting gap is $\sim$$14$pp. The cross-model single-frame probe is confounded (see text).}
\label{tab:channel}
\begin{tabular}{lc}
\toprule
Retrieval query for the \emph{true} state & R@5 \\
\midrule
Canonical GT string (semantic oracle) & 53.0 \\
LLM paraphrase of GT (content fixed) & 29.7 \\
Non-Gemini VLM single-frame description$^\dagger$ & 3.2 \\
\midrule
\emph{Best LFTR (learned predictor)} & \emph{18.6} \\
\bottomrule
\end{tabular}
\end{table}

\FloatBarrier
\section{Qualitative Retrieval --- Full Ranked List}
\label{app:qual}
\begin{figure*}[t]\centering
\includegraphics[width=\textwidth]{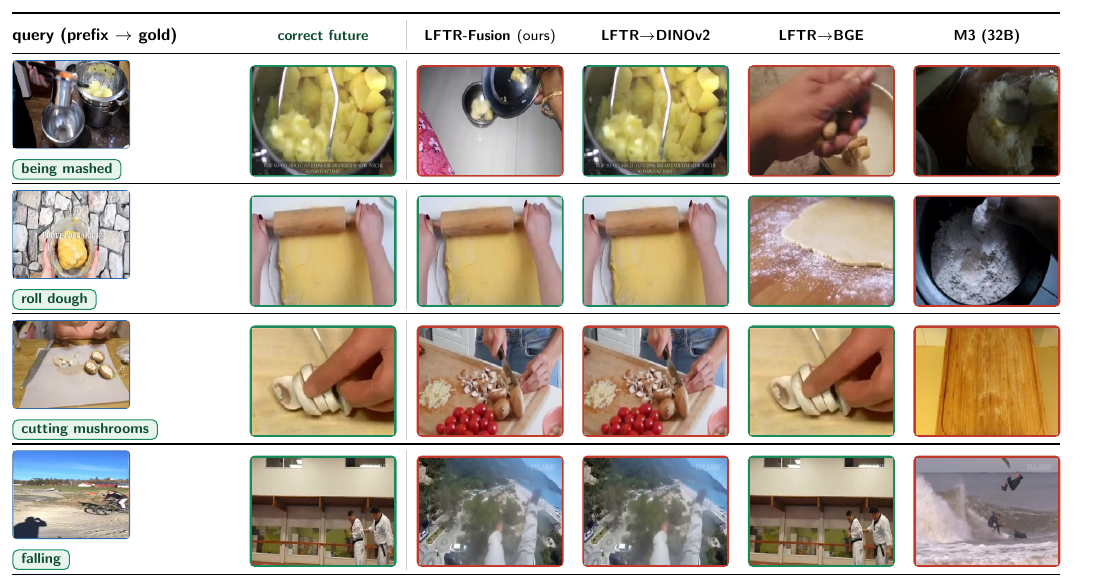}
\caption{\textbf{Top-1 retrieved future state per method}, one query per row (observed prefix and its gold
future on the left; each cell the single top-ranked retrieval, \textcolor{green!55!black}{green} if its state
matches the gold future, \textcolor{red!62!black}{red} otherwise). The zero-shot 32B MLLM (M3) never places
the correct future at rank~1; the learned cross-space retrievers recover it on several queries. Top-1 is a
strict view --- the full top-5 ranking (where fusion leads) is in Fig.~\ref{fig:top5}.}
\label{fig:qual}
\end{figure*}

\begin{figure*}[t]\centering
\includegraphics[width=\textwidth]{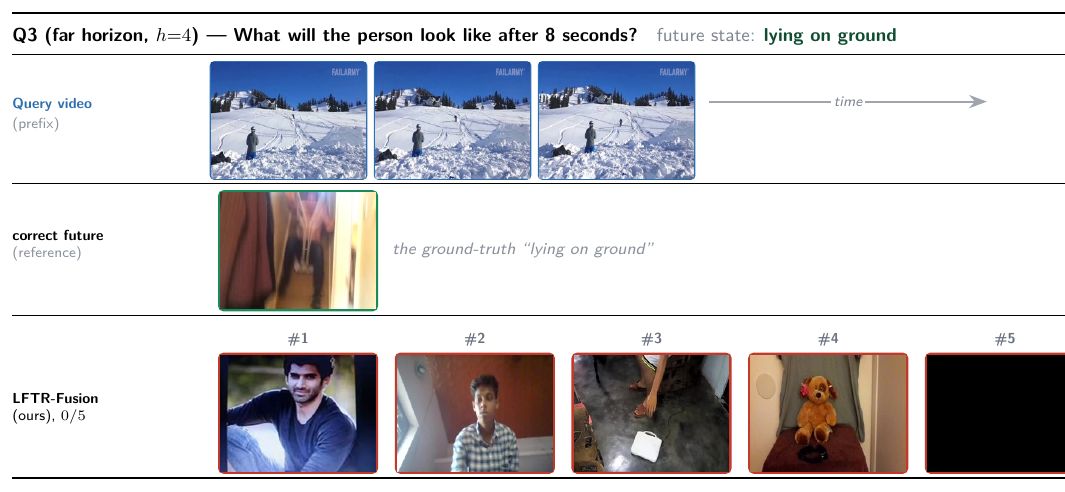}
\caption{\textbf{Far-horizon failure.} A long-horizon ($h{=}4$) query where fusion is wrong on all five
retrievals: the correct future (\textcolor{green!55!black}{green} reference, \emph{lying on ground}) versus
LFTR-fusion's top-5 (\textcolor{red!62!black}{red}, $0/5$). At far horizons the future is genuinely
multi-modal, so even fusion returns plausible-but-wrong states --- the prediction bottleneck, made concrete.}
\label{fig:farfail}
\end{figure*}

\begin{figure*}[t]\centering
\includegraphics[width=\textwidth]{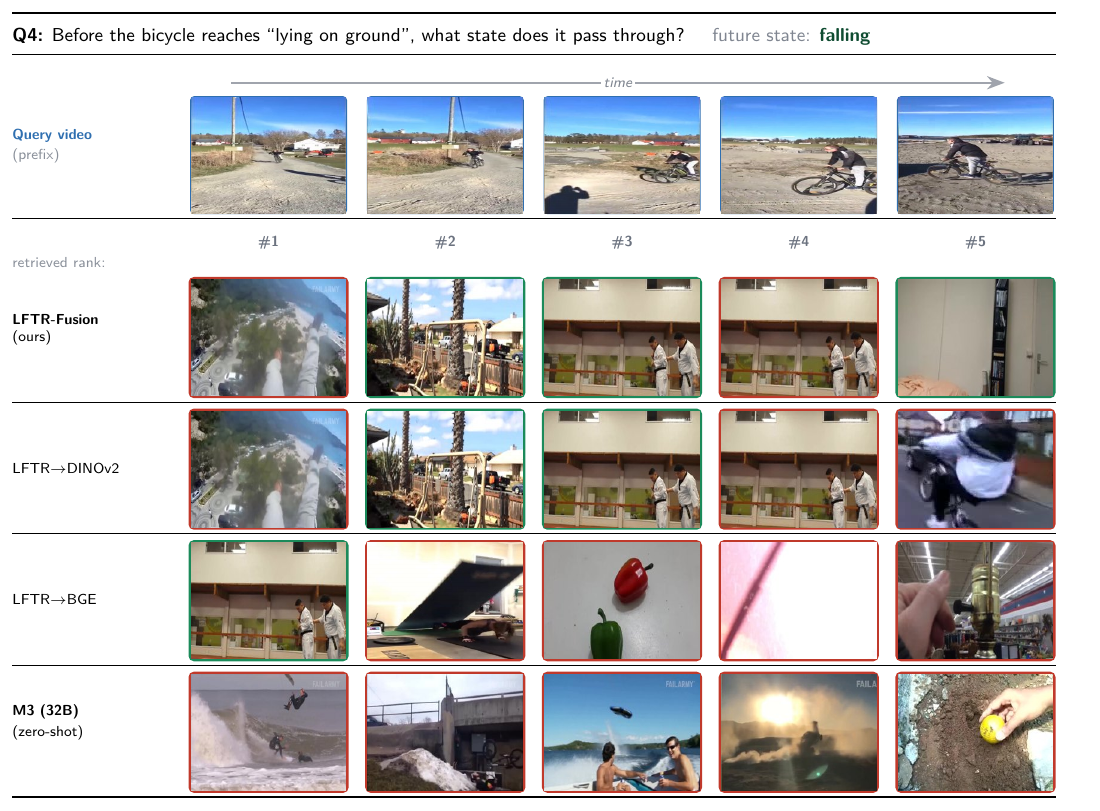}
\caption{\textbf{Top-5 retrievals per method} for one query (Q4; future state \emph{falling}). The observed
prefix (blue, ordered left-to-right in time) is shown on top; below, each method's top-5 retrieved future
segments carry a \textcolor{green!55!black}{green} border when the retrieved state matches the ground-truth
future and a \textcolor{red!62!black}{red} one otherwise. Cross-space fusion (ours) places the most correct
futures in its top-5 ($3/5$), degrading through LFTR$\rightarrow$DINOv2 ($2/5$) and LFTR$\rightarrow$BGE
($1/5$) to the zero-shot 32B MLLM (M3, $0/5$).}
\label{fig:top5}
\end{figure*}

This section collects the qualitative retrieval examples referenced above: for real queries, the actual
future segments each method retrieves, with a \textcolor{green!55!black}{green} border when the retrieved
state matches the gold future and \textcolor{red!62!black}{red} otherwise. We show the top-1 retrieval
across methods (Fig.~\ref{fig:qual}), far-horizon failures (Fig.~\ref{fig:farfail}), the full top-5 ranked
lists (Fig.~\ref{fig:top5}), and per-question-type and per-method breakdowns
(Figs.~\ref{fig:btQ1}--\ref{fig:pmM3}).


Fig.~\ref{fig:top5} extends the top-1 comparison (Fig.~\ref{fig:qual}) to the top-5 ranked list of each
method. Cross-space fusion places the most correct instances (green) in its top-5, while the naive and MLLM
baselines return mostly hard-negative states (red) --- the fine-grained-discrimination failure that
hard-negative training targets (App.~\ref{app:hardneg}). Absolute counts are modest (R@5 is $\le$$17$ even
for fusion), reflecting the intrinsic difficulty of the task. Fig.~\ref{fig:farfail} isolates the
\emph{far-horizon} ($h{\ge}3$) failures: the multi-modal future drives even fusion to a plausible-but-wrong
adjacent state, making the prediction bottleneck concrete.

\subsection{Per-question-type retrieval examples}
\label{app:bytype}
\begin{figure*}[t]\centering
\includegraphics[width=0.86\textwidth]{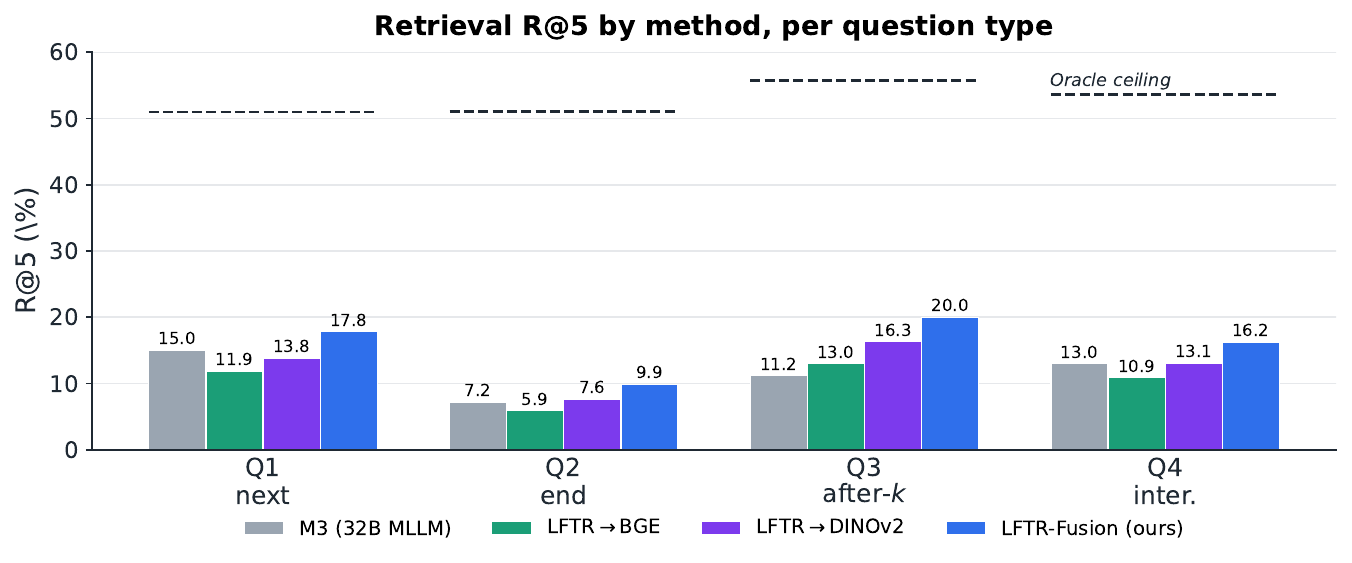}
\caption{\textbf{Retrieval R@5 by method, per question type} (aggregate over the test set). Cross-space
fusion (ours) leads on every question type; the dashed line is the Oracle-BGE ceiling, far above all methods
--- the prediction headroom the benchmark exposes. Q2 (end state) is hardest for every method.}
\label{fig:qtypesummary}
\end{figure*}

\begin{figure*}[t]\centering
\includegraphics[width=\textwidth]{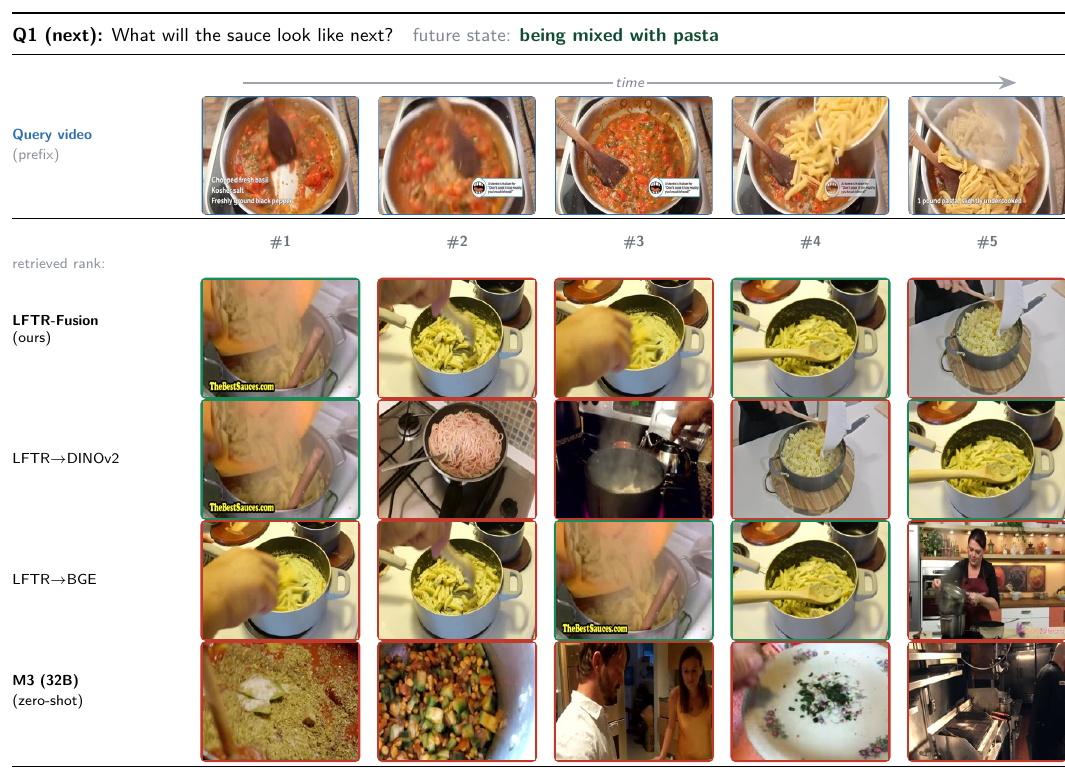}
\caption{\textbf{Q1 (next state) --- top-5 retrievals per method.}}
\label{fig:btQ1}
\end{figure*}

\begin{figure*}[t]\centering
\includegraphics[width=\textwidth]{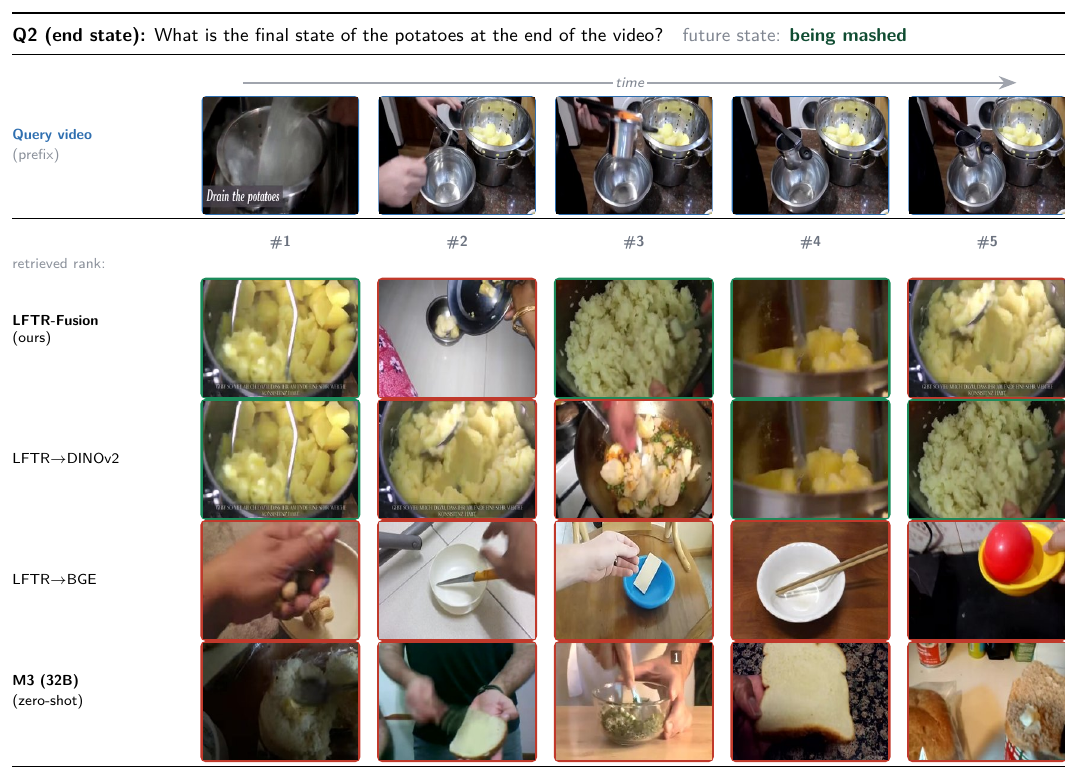}
\caption{\textbf{Q2 (end state) --- top-5 retrievals per method.}}
\label{fig:btQ2}
\end{figure*}

\begin{figure*}[t]\centering
\includegraphics[width=\textwidth]{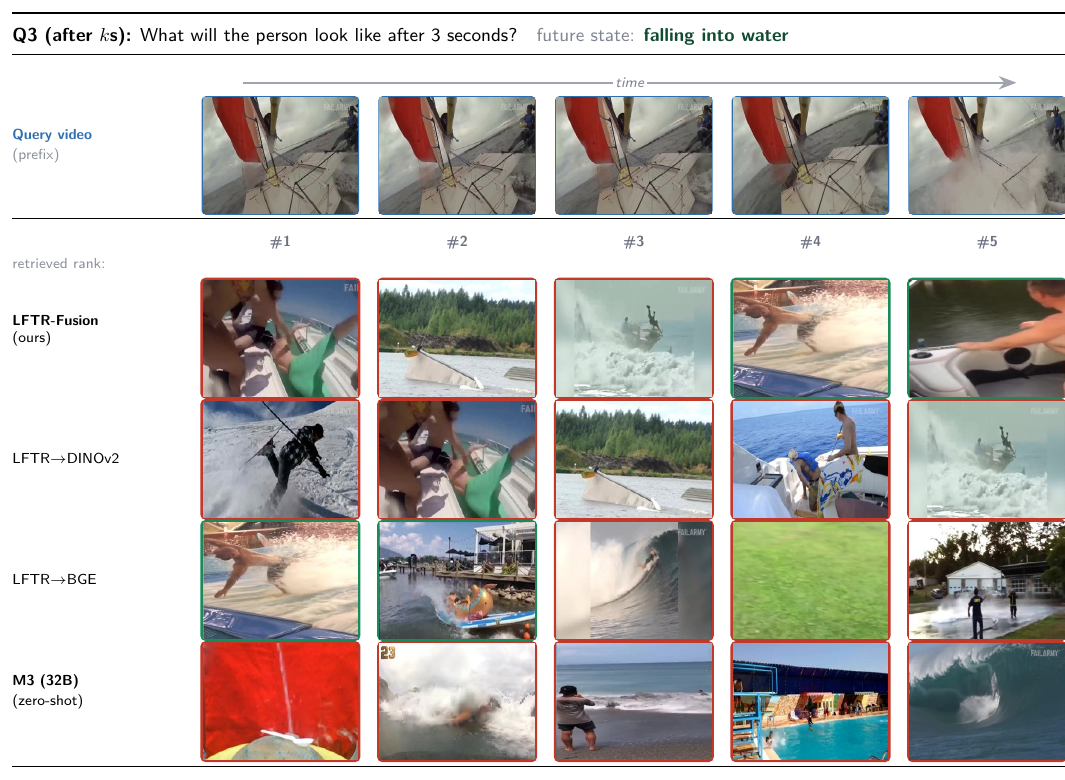}
\caption{\textbf{Q3 (state after $k$ seconds) --- top-5 retrievals per method.}}
\label{fig:btQ3}
\end{figure*}

\begin{figure*}[t]\centering
\includegraphics[width=\textwidth]{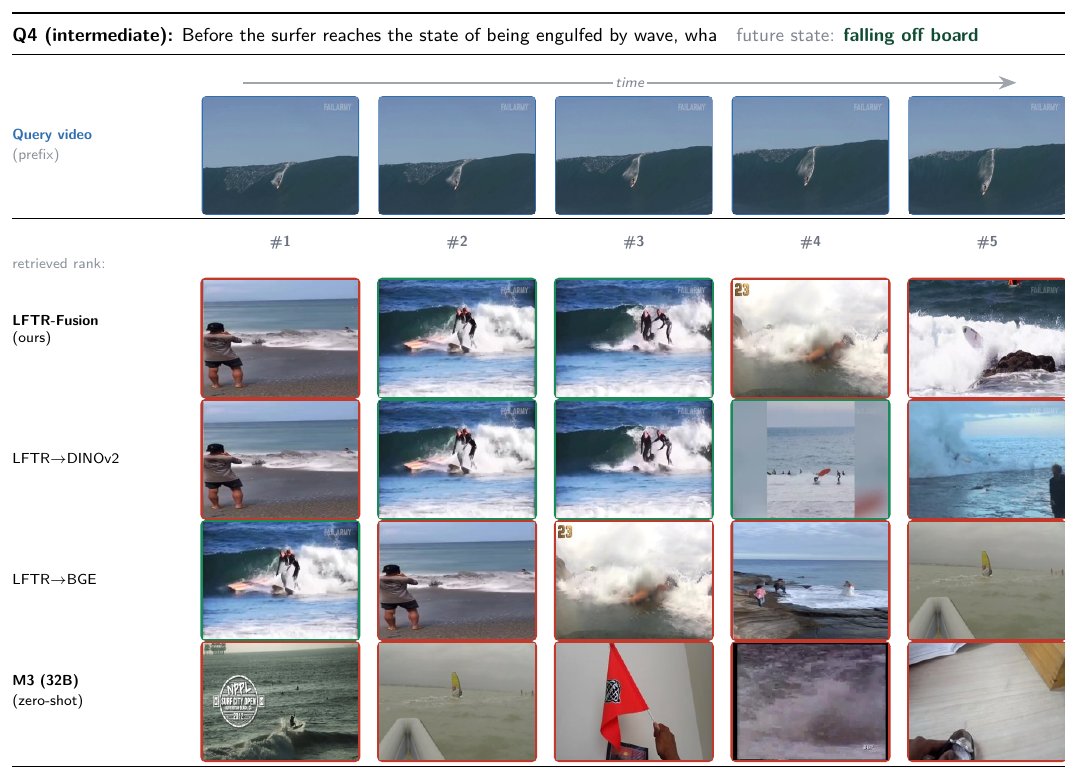}
\caption{\textbf{Q4 (intermediate state) --- top-5 retrievals per method.}}
\label{fig:btQ4}
\end{figure*}

Figures~\ref{fig:btQ1}--\ref{fig:btQ4} give one top-5 retrieval example \emph{per question type} (Q1--Q4),
in the same format as Fig.~\ref{fig:top5}: the observed prefix (blue, ordered in time) on top and, per
method, the top-5 retrieved future segments (\textcolor{green!55!black}{green} correct,
\textcolor{red!62!black}{red} wrong). Across types, cross-space fusion places more correct futures near the
top of the list than the semantic-only, visual-only, and zero-shot 32B MLLM (M3) baselines. Figure~\ref{fig:qtypesummary}
first summarizes the \emph{aggregate} R@5 of each method on each type.


\subsection{Per-method retrieval across question types}
\label{app:permethod}
\begin{figure*}[t]\centering
\includegraphics[width=\textwidth]{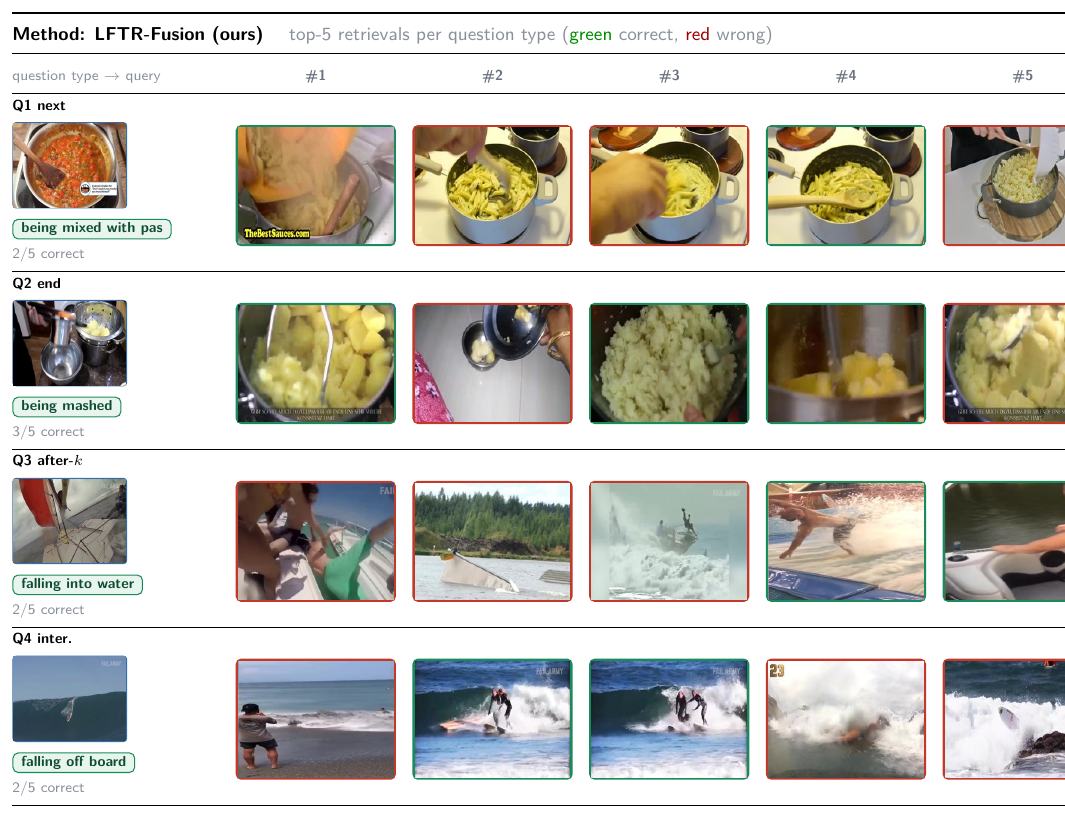}
\caption{\textbf{LFTR-Fusion (ours)} --- top-5 retrievals across question types.}
\label{fig:pmFusion}
\end{figure*}

\begin{figure*}[t]\centering
\includegraphics[width=\textwidth]{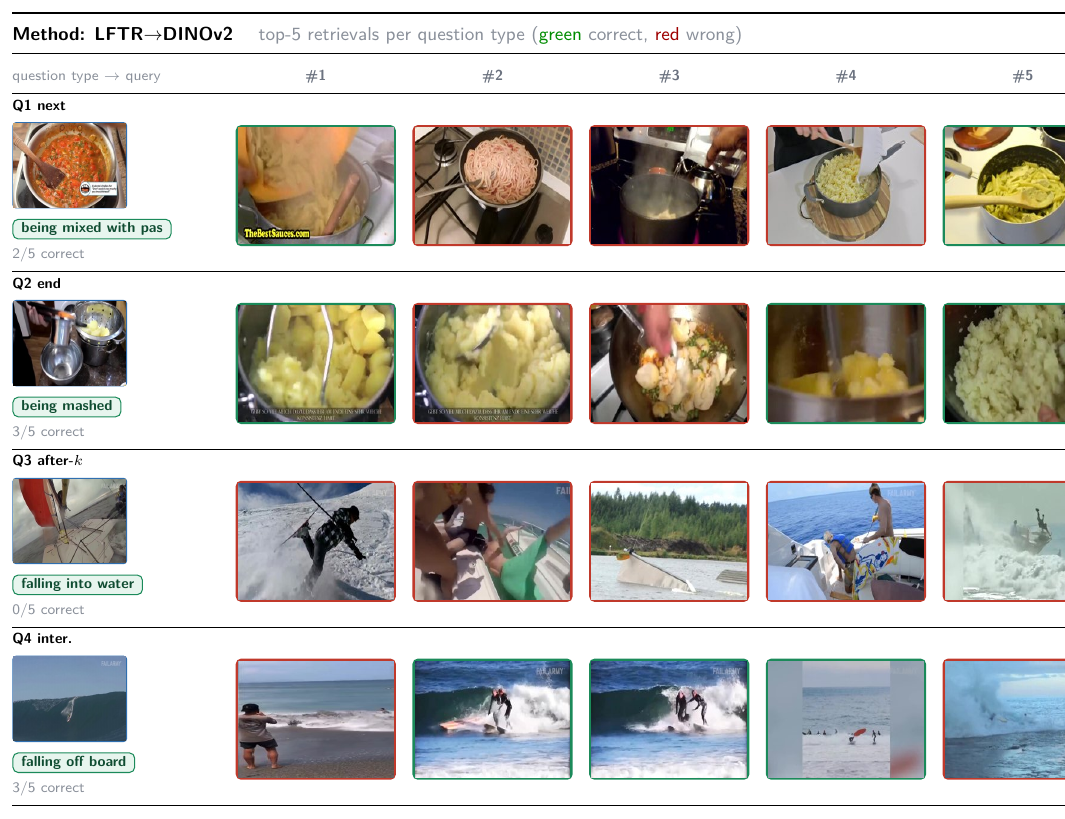}
\caption{\textbf{LFTR$\rightarrow$DINOv2} --- top-5 retrievals across question types.}
\label{fig:pmDino}
\end{figure*}

\begin{figure*}[t]\centering
\includegraphics[width=\textwidth]{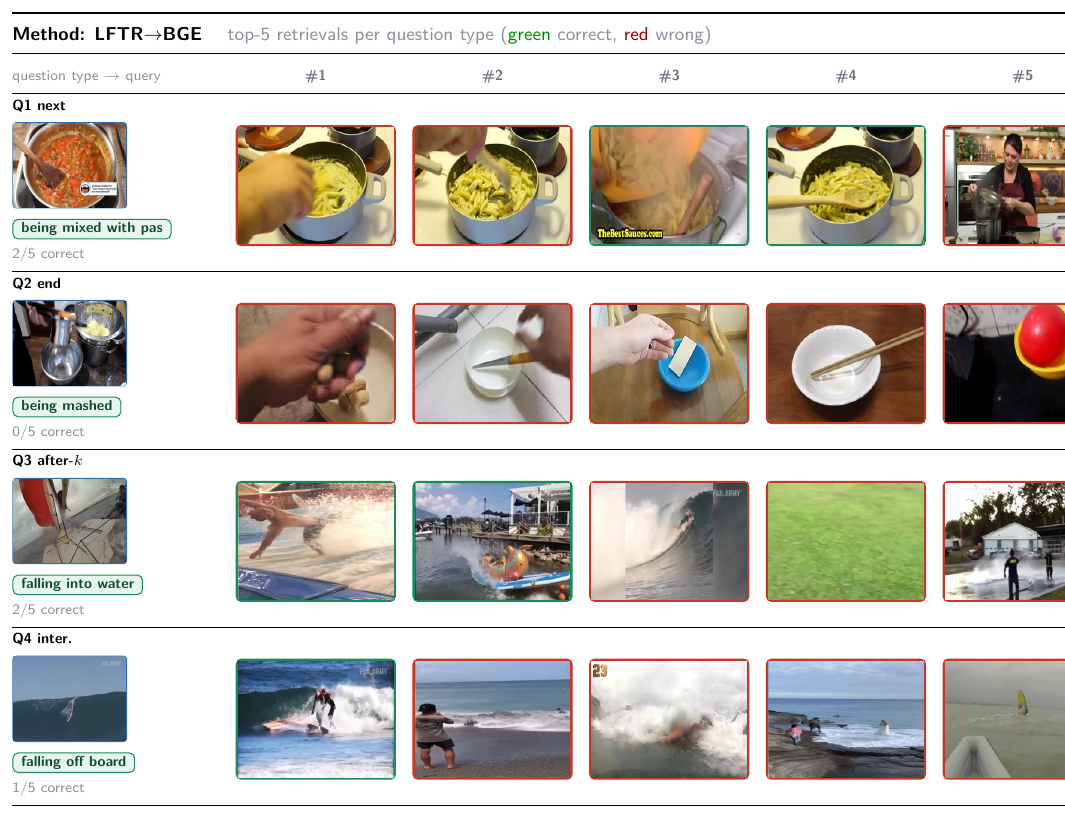}
\caption{\textbf{LFTR$\rightarrow$BGE} --- top-5 retrievals across question types.}
\label{fig:pmBge}
\end{figure*}

\begin{figure*}[t]\centering
\includegraphics[width=\textwidth]{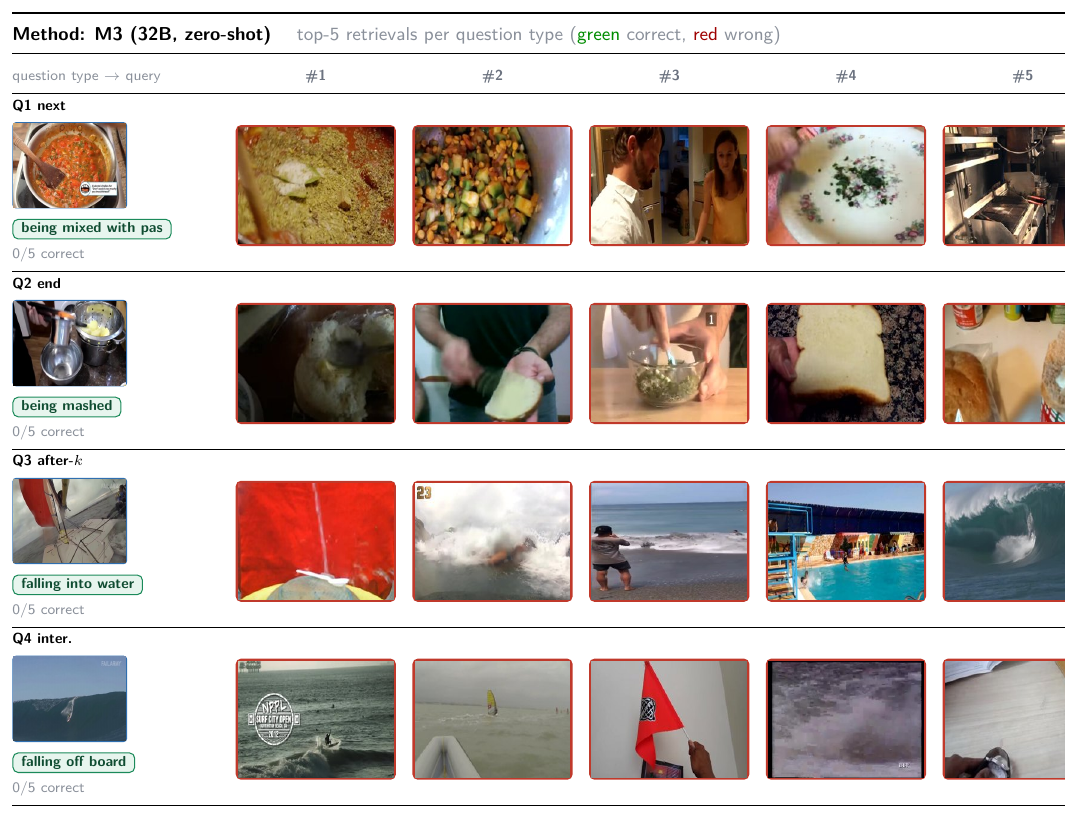}
\caption{\textbf{M3 (zero-shot 32B MLLM)} --- top-5 retrievals across question types.}
\label{fig:pmM3}
\end{figure*}

The complementary view: Figures~\ref{fig:pmFusion}--\ref{fig:pmM3} give \emph{one figure per method}, each
with one row per question type (Q1--Q4), showing that method's top-5 retrievals and how many are correct.
Cross-space fusion (Fig.~\ref{fig:pmFusion}) is consistently strongest; the zero-shot 32B MLLM
(Fig.~\ref{fig:pmM3}) is weakest across every type.


\end{document}